\documentclass[10pt,twocolumn,letterpaper]{article}

\usepackage[pagenumbers]{wacv} %

\usepackage[most]{tcolorbox}
\usepackage{cuted}
\usepackage{booktabs}
\usepackage{adjustbox}
\usepackage{tabularx}
\usepackage{svg}
\usepackage{array}
\newcolumntype{Y}{>{\raggedright\arraybackslash}X}

\usepackage[normalem]{ulem}
\useunder{\uline}{\ul}{}

\newcommand{\bn}{(B)RAVEn\xspace}

\newcommand{\ourname}{BabelFake\xspace}
\newcommand{\DF}{DeepFake}

\definecolor{wacvblue}{rgb}{0.21,0.49,0.74}
\usepackage[pagebackref,breaklinks,colorlinks,allcolors=wacvblue]{hyperref}

\def\wacvPaperID{177} %
\def\confName{WACV}
\def\confYear{2027}

\title{BabelFake: A Multilingual Audio-Visual DeepFake Benchmark}

\author{
Carlotta Segna \qquad Joel Tschesche \qquad Anna Rohrbach\\
TU Darmstadt \& Hessian.AI, Germany\\
}

\begin{document}
\maketitle
\begin{abstract}

Reliable and practical audio-visual DeepFake detection requires benchmarks that reflect diverse linguistic contexts and modern data synthesis pipelines for visual as well as audio manipulations. However, existing datasets predominantly contain footage of English-speakers, often include outdated manipulation types, or overlook the audio modality. Further, many datasets feature individuals who did \emph{not} consent to be used in DeepFake creation. 
We introduce \emph{\ourname}, a multilingual audio-visual DeepFake benchmark recorded with consenting participants. \ourname contains 399k clips (1,323 hours) from 496 individuals spanning five languages (English, German, Italian, French, Spanish). Our modular data generation pipeline pairs 11 modern video manipulation methods with 4 voice cloning engines, distinguishing visual-only (face swapping) and joint audio-visual manipulations (lip synchronization and portrait animation). 
By benchmarking state-of-the-art detectors, we show that detection difficulty depends on the audio-visual generation pairing, with substantial performance degradation when authentic audio is preserved. Cross-language/demographic evaluation reveals sensitivity varying across detector architectures and training data, while human evaluation reveals that perceived realism and machine-detection difficulty do not necessarily align.  

\end{abstract}
    
\section{Introduction}
\label{sec:intro}

The rapid evolution of generative models has dramatically increased the realism and accessibility of synthetic media. Modern frameworks can seamlessly clone speech~\cite{fishaudio, chatterbox, omnivoice}, reenact facial dynamics~\cite{EmoTalk, Face2Face}, or entirely replace identities in video~\cite{SimSwap, HifiFace}. While enabling beneficial applications, they also pose severe risks for disinformation and digital impersonation~\cite{intro_df1, intro_df2, intro_df3}, making robust \DF\ detection a matter of priority.

Progress in \DF\ detection is driven by benchmarks. Early visual \DF\ datasets, such as FaceForensics++~\cite{FaceForensics}, Celeb-DF~\cite{celeb-df}, KoDF~\cite{KoDF}, and DFDC~\cite{DFDC2020}, significantly advanced research in detecting visual manipulation. More recently, audio-visual datasets such as FakeAVCeleb~\cite{FakeAVCeleb}, LAV-DF~\cite{lav-df}, and DeepSpeak~\cite{DeepSpeak} have advanced the development of multimodal detection approaches that jointly analyze audio and visual information. 
Despite these advances, existing datasets present several critical limitations: (i) \emph{Monolingual Bias}: most datasets remain predominantly English-centric, and the emerging multilingual datasets~\cite{ILLUSION, mavos} originate from uncontrolled web-scraped media; (ii) \emph{Outdated Manipulation Types}: although manipulation capabilities improved during the four-year span from FakeAVCeleb to DeepSpeak, the latter still retains some older manipulation types, limiting its alignment with contemporary deepfake generation frameworks; (iii) \emph{Overlooked Audio}: existing audio-visual benchmarks are heavily skewed towards visual manipulations. While they provide extensive variations of face swapping and lip synchronization, they include few audio manipulations; (iv) \emph{Consent}: the vast majority of \DF\ benchmarks scrape videos of public figures or online users without explicit consent, raising privacy and data-use concerns. 

\begin{table*}[t!]
\centering
\resizebox{\textwidth}{!}{
\begin{tabular}{lccccccccccc}
\hline
\multicolumn{1}{c}{\textbf{Dataset}}       & \textbf{Year} & \textbf{\# Lang.} & \textbf{English \%} & \textbf{Consent/} & \textbf{\# Manipulations} & \multicolumn{3}{c}{\textbf{Video Manipulations}}                                              & \textbf{Audio} & \multicolumn{2}{c}{\textbf{Samples}}                \\ \cline{7-9} \cline{11-12} 
\multicolumn{1}{c}{}                       &               &                   &                     & \textbf{Ethics Approval}    &                           & \textbf{Faceswap}              & \textbf{Lipsync}               & \textbf{Portrait Animation} &                & \textbf{Real Samples}       & \textbf{Fake Samples} \\ \hline
PolyGlotFake~\cite{PolyGlotFake}                               & 2024          & 7                 & 16.7\%              & N/N               & 7                        & \multicolumn{1}{c|}{--}        & \multicolumn{1}{c|}{\checkmark} & --                          & \checkmark      & \multicolumn{1}{c|}{766}    & 14,472                \\
ILLUSION - Set D~\cite{ILLUSION}                           & 2025          & 26                & UnK                  & N/\checkmark       & UnK                        & \multicolumn{1}{c|}{UnK}        & \multicolumn{1}{c|}{UnK}        & UnK                          & UnK             & \multicolumn{1}{c|}{100}    & 125                   \\
MAVOS-DD~\cite{mavos}                                   & 2025          & 8                 & 22.6\%              & N/N               & 7                         & \multicolumn{1}{c|}{\checkmark} & \multicolumn{1}{c|}{--} & \multicolumn{1}{c}{\checkmark}                       & --             & \multicolumn{1}{c|}{25,195} & 35,169                \\
\ourname(Ours) & 2026          & 5                 & 20.2\%        & \checkmark/\checkmark              & 15                        & \multicolumn{1}{c|}{\checkmark} & \multicolumn{1}{c|}{\checkmark} & \checkmark                   & \checkmark      & \multicolumn{1}{c|}{20,117} & 379,582               \\ \hline
\end{tabular}
}
\caption{Comparison between \ourname and other recent multilingual DeepFake datasets. For ILLUSION, only set D (video subset) is included in the comparison, as it is the only one considering the multilingual scenario, UnK means unknown.}
\label{tab:dataset_comparison}
\end{table*}

To overcome these limitations, we introduce \textbf{\ourname}, a multilingual audio-visual \DF\ benchmark recorded with consenting participants and approved by our Institutional Review Board (IRB, an ethics committee). Captured via a controlled webcam protocol, the dataset reflects realistic remote communication scenarios. It comprises recordings from speakers in five languages (English, German, French, Italian, and Spanish) who explicitly consent to the collection, processing, and release of their data for academic research. All recordings were captured on personal devices in unconstrained home environments from diverse geographic locations, ensuring natural variability while maintaining consistent acquisition constraints. To capture realistic conversational behavior, our protocol incorporates a structured task hierarchy spanning phonetically rich reading, quick-fire Q\&A, spontaneous open-ended storytelling, and action- or gesture-based prompts. %

Next, we implement a data generation protocol that addresses audio modality under-representation. We combine 11 state-of-the-art video manipulation methods with 4 recent multilingual voice-cloning engines. This design covers visual-only generation (Face Swapping), and joint audio-visual manipulations (Lip Synchronization and Portrait Animation) conditioned on both pristine and synthetic audio. 

We benchmark several state-of-the-art DeepFake detectors (SpeechForensics~\cite{speechforensics}, AVH-Align~\cite{AVH-Align}, \bn~\cite{braven}) on \ourname. Our evaluation reveals that existing detectors suffer performance drops when visual manipulations are paired with authentic audio, hinting at over-reliance on cross-modal synchronization cues in multimodal models. While multimodal models generally outperform their unimodal counterparts, their advantage varies considerably across audio-video generation combinations. Finally, multilingual evaluation shows no language is consistently harder to detect, but highlights language sensitivity in the detector architecture and training data. 

Our contributions can be summarized as follows: (i) \emph{Ethically Sourced Benchmark:} we release \ourname, an IRB-approved multilingual \DF\ detection dataset featuring recordings with consenting participants across 5 languages, ensuring diverse linguistic/demographic representation and responsible data use; (ii) \emph{Modality-Disentangled Generation Pipeline}: we introduce a modular generation protocol pairing 11 video manipulation engines with 4 voice-cloning models, covering video-only manipulations alongside audio-visual DeepFakes under both authentic and synthesized speech conditions; (iii) \emph{Diagnostic Insights:} we present a comprehensive evaluation of state-of-the-art multimodal detectors, exposing their blind spots to authentic audio and analyzing their cross-lingual robustness. \ourname also enables analysis across demographic groups. 

\section{Related Work}
\label{sec:related_work}

\textbf{Audio-Visual DeepFake Datasets.}
The availability of large-scale datasets plays a crucial role in advancing audio-visual \DF\ detection. Early datasets, such as DFDC~\cite{DFDC2020}, primarily focus on visual manipulation and large-scale face-swapping scenarios. To support multimodal detection, FakeAVCeleb~\cite{FakeAVCeleb} introduces synchronized manipulations on both facial and speech, enabling the study of audio-visual inconsistency arising from jointly manipulated content. More recently, DeepSpeak~\cite{DeepSpeak} expands the diversity of manipulation methods and generation pipelines, providing more challenging benchmarks for multimodal \DF\ detection. %
Despite these advances, existing audio-visual datasets are primarily designed to improve manipulation diversity and scale, while offering limited support for systematic analysis across languages and modality configurations. These aspects are essential for understanding the robustness and generalization capabilities of current \DF\ detectors. 

\textbf{Multilingual DeepFake Datasets.}
The growing availability of multilingual generative models has motivated the development of \DF\ benchmarks that extend beyond English. Early efforts include KoDF~\cite{KoDF}, which introduces Korean-speaking subjects for visual \DF\ detection. More recently, multilingual audio-visual benchmarks such as PolyGlotFake~\cite{PolyGlotFake} and MAVOS-DD~\cite{mavos} have expanded the linguistic diversity of \DF\ datasets by including multiple languages, subjects, and generation methods. ILLUSION~\cite{ILLUSION} further introduces a large-scale benchmark spanning image, audio, video, and audio-visual DeepFakes. However, to the best of our knowledge, among its subsets, \emph{only subset D supports the multilingual scenario}, while the remaining subsets primarily focus on large-scale multimodal generation in English. Consequently, only Set D is considered in our comparison.\footnote{We were not able to access the data to confirm all the details.}

Although these datasets represent important steps towards multilingual evaluation, %
as shown in Tab.~\ref{tab:dataset_comparison}, \ourname is the first multilingual audio-visual \DF\ benchmark to combine consented and controlled data acquisition with multiple manipulation modalities, including face swapping, lip-syncing, avatar generation, and audio DeepFakes, thereby enabling a comprehensive evaluation of detector robustness across languages and manipulation types.

\section{\ourname: Real Data Collection}
\label{sec:dataset_design}

In this section we outline our scheme for real data recording, with the goal to later facilitate multilingual multimodal DeepFake data generation (described in Sec.~\ref{sec:generation_pipeline}).

\subsection{Design Goals}
The design of \ourname was guided by three primary objectives: ethical data collection, multilingual coverage, and realistic data acquisition conditions. First, unlike many existing \DF\ benchmarks that rely on publicly available online footage, we adopt a consent-first acquisition strategy, in which participants explicitly agree to the collection, processing, and public release of their recordings for research purposes. Additionally, the data acquisition protocol was reviewed and approved by our institutional ethics committee. This approach addresses the growing concerns regarding privacy, data ownership, and the ethical use of biometric information in \DF\ research~\cite{consent1, consent2}. 

Second, the dataset is designed to support multilingual \DF\ detection. While modern generative approaches can produce DeepFakes in many languages, most existing benchmarks remain English-centric. To address this limitation, \ourname includes recordings in five languages - English (en), German (de), Italian (it), French (fr), and Spanish (es) - enabling the study of language-specific effects and cross-language generalization. Moreover, we do not only capture distinct languages, but, consequently, distinct geographic locations and ethnic groups (see Fig.~\ref{fig:nationality_top10}).

Finally, the dataset aims to capture realistic webcam-based interactions. As remote interactions become increasingly common, these settings represent an important real-world deployment scenario for \DF\ detection systems. All recordings were collected using participants' personal devices in an unconstrained home environment, while following standardized recording protocol designed to ensure good audio and video quality. 

\subsection{Participant Recruitment and Consent}
Participants were recruited through the Prolific\footnote{https://www.prolific.com/} crowdsourcing platform and were financially compensated based on the estimated study duration (ca. 40 minutes), following Prolific's recommended hourly compensation guidelines (\pounds9/hour). Data collection was conducted in four stages: an initial pilot study, followed by three collection phases. Each language was collected through a dedicated study, with the target language as the primary criterion for recruitment. To avoid duplicate participants and ensure unique identities in the dataset, participants who completed any collection study were excluded from all subsequent studies. For more details, see ~\cref{app:real_data}.
Before starting the recording phase, participants reviewed a detailed consent form (see~\cref{sec:consent_form}). The form explains the study's purpose, data recording and storing, public dataset release, potential risks, and how to withdraw consent. %

\subsection{Recording Protocol \& Speech Content}
We collect the data through a custom web-based platform; participants complete the task on their own devices in their everyday environments. %
Our recording protocol combines scripted reading, spontaneous Q\&A interactions, open responses, emotional speech prompts, and action-based instructions, inspired by DeepSpeak~\cite{DeepSpeak}. Jointly, these tasks capture a wide range of speech patterns, facial expressions, head poses, gestures, and natural behaviors typical of everyday video calls. 

For \emph{voice cloning} tasks, participants complete a set of standardized speech tasks. These tasks provide broad phonetic coverage and allow direct comparison among speakers of the same language. Specifically, participants record a \emph{language-specific} version of ``The North Wind and the Sun''~\cite{aesop_northwind} and a set of phonetically balanced sentences. 

To capture natural speaking behavior, participants complete \emph{question-answering} (e.g., concise answers) and \emph{open-ended} (e.g., spontaneous storytelling) responses to encourage natural habits like pauses, hesitations, filler words, varying speech rates, and spontaneous gestures. 

We also include \emph{emotional reading prompts} to add variety to both audio and video. These prompts capture changes in tone, intonation, speaking rate, facial expressions, and emotional behavior across happy, sad, and neutral speech. 

Finally, participants follow \emph{action-based instructions} that mimic behaviors common in real-world video calls. These tasks trigger movements, gaze shifts, facial expressions, partial face obstructions, and speech-related gestures, which provide extra visual variety needed for audio-video \DF\ generation and detection.

We choose text prompts that maximize linguistic diversity and phonetic coverage, while ensuring content safety. Further details on prompt sources, filtering procedures, and prompt selection are presented in ~\cref{app:prompt_source}; the text prompts are given in~\cref{app:textual_prompts}.

\section{\ourname: DeepFake Generation Pipeline}
\label{sec:generation_pipeline}
Starting from the pristine recordings described in ~\cref{sec:dataset_design}, we generate manipulated samples via a modular pipeline combining preprocessing, controlled identity pairing, multilingual voice cloning, and multiple types of video manipulation (both visual-only and audio-visual).

\subsection{Preprocessing Pipeline}
Raw recordings collected via our web acquisition platform are standardized to MP4 (H.264 video, AAC audio at 48kHz). To support text-driven audio synthesis, we transcribe the authentic recordings using Whisper Large~\cite{whisper} and post-process the transcripts with Gemini 2.5~\cite{gemini} to remove transcription errors and ensure grammatical flow for text-to-speech (TTS) voice synthesis input.

For generative pipelines based on cloned speech, audio tracks are resampled to 16kHz. To prevent temporal-looping artifacts or unrealistic static padding, we enforce strict audio-visual duration alignment by trimming the source video when it exceeds the driving audio by more than 0.5s and discarding samples where the driving audio exceeds the video length. Full encoding, frame rate handling, and alignment thresholds are detailed in~\cref{app:preprocessing}.

\subsection{Identity Pairing}
\label{sec:identity_pairing}
To generate realistic identity manipulations,~\ourname employs a controlled demographic pairing strategy for both facial and vocal identities. For face-based manipulations, each target participant is paired with a source using a hierarchical demographic matcher. The demographic matcher is based on the language, sex, age bucket (18-35, 35-50, 50+), and ethnicity.  When no candidate matches the full criteria, the constraints are progressively relaxed by first removing ethnicity, then age, and finally sex, while language is always preserved, until a match is found.
For each selected face source, the reference image is selected on the ``best eye pose'' (more details in Sec.~\ref{app:best_eye_pose}), using the question-answering videos as the starting point. %

Regarding the fake audio generation, the target provides the video identity and transcript, while the source voice is the reference speech used for voice cloning. By separating facial and vocal identity, ~\ourname supports manipulations in which the target face and voice source are composed of two different participants. 

\subsection{DeepFake Generation}
We generate manipulated videos using several state-of-the-art DeepFake generation methods, covering $3$ types of manipulation: \emph{face swapping, lip synchronization, and portrait animation}. Implementation details and descriptions of all the audio and video manipulation models are provided in ~\Cref{app:generation_techniques}. Here, we summarize the key characteristics of the employed methods and respective generation settings; note, only three methods predate 2025, all others are very recent.

\textbf{Face Swapping.} We use CanonSwap~\cite{canonswap} (2025), DeepLive Cam~\cite{deeplivecam} (2025), HiFiFace~\cite{HifiFace} (2021), and InSwapper~\cite{inswapper} (2024). For each target video, a face is selected by randomly picking a source matching identity following the rules above (\cref{sec:identity_pairing}). Face-swapping methods preserve the target video's motion, facial expression, and original audio while replacing the visible facial identity. 

\textbf{Speech Synthesis \& Lip Synchronization}. Synthetic audio is generated by conditioning the target transcript on the assigned source voice donor. We use $4$ multilingual text-to-speech methods: Qwen3-TTS~\cite{Qwen} (2026), FishAudio2~\cite{fishaudio} (2026), OmniVoice~\cite{omnivoice} (2026), and ChatterBox~\cite{chatterbox} (2025). The synthetic audio is used to drive lip synchronization and a subset of portrait animation methods. 
Using KeySync~\cite{keysync} (2025), LatentSync~\cite{latentsync} (2024), and MuseTalk~\cite{musetalk} (2025), we alter the mouth dynamics of a source video to match a target audio (real or synthetic). 
 
\textbf{Portrait Animation.} We use AvatarForcing~\cite{avatarforcing} (2026), PersonaLive~\cite{personalive} (2025), HunyuanPortrait~\cite{hunyuanportrait} (2025), and UniAVGen~\cite{uniavgen} (2025) for the last category of manipulations. In this case, starting from a static image of a target identity, animation is driven by audio (AvatarForcing, UniAVGen) or by video (HunyuanPortrait, PersonaLive).

\section{\ourname Analysis}

\begin{figure}[t]
    \centering
    \includegraphics[width=\linewidth]{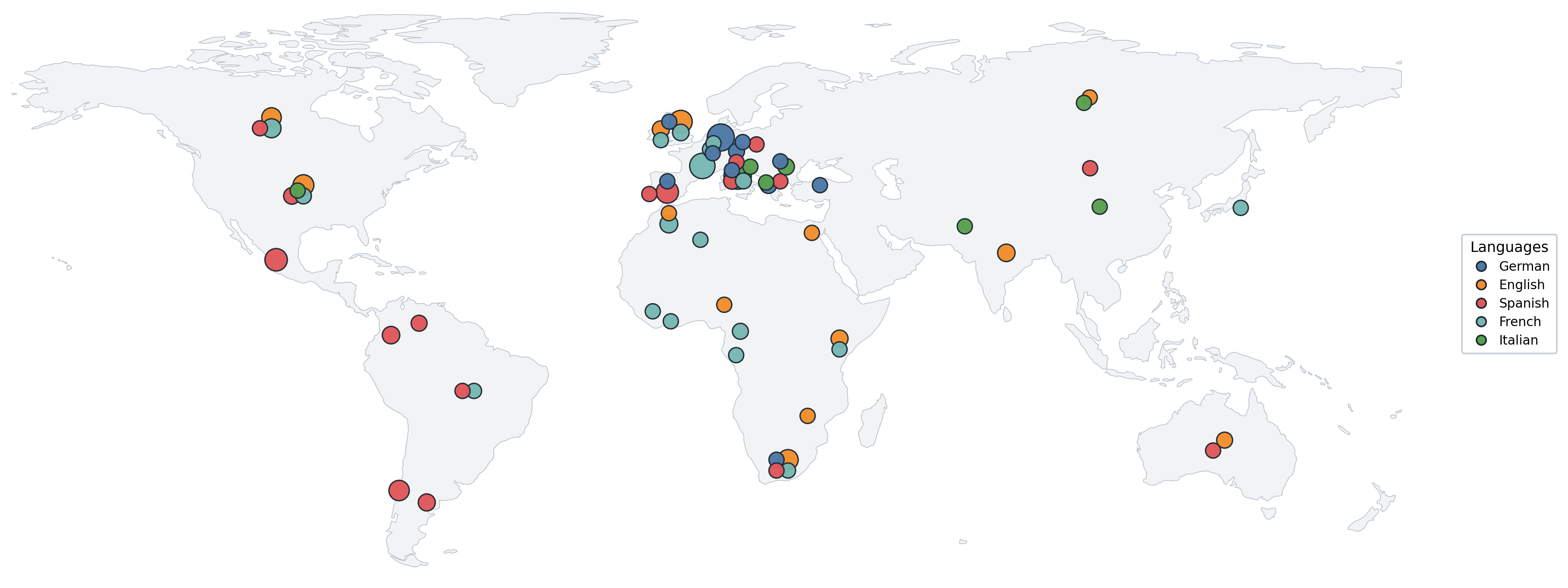}
    \vspace{-5mm}
    \caption{Geographic diversity of \ourname. Each marker corresponds to one or more participants, with marker size proportional to the number of subjects collected at that location. Colors indicate the language in which the participant was recorded. The figure illustrates the worldwide geographic diversity of the benchmark and shows that each language subset contains participants from multiple countries rather than a single geographic region.}
    \label{fig:nationality_top10}
\end{figure}

\begin{table}[t]
\centering
\resizebox{\columnwidth}{!}{%
\begin{tabular}{llllllll}
\hline
Language & Participants & \begin{tabular}[c]{@{}l@{}}Real \\ Clips\end{tabular} & \begin{tabular}[c]{@{}l@{}}Fake \\ Clips\end{tabular} & \begin{tabular}[c]{@{}l@{}}Total \\ Clips\end{tabular} & \begin{tabular}[c]{@{}l@{}}Real \\ Hours\end{tabular} & \begin{tabular}[c]{@{}l@{}}Fake \\ Hours\end{tabular} & \begin{tabular}[c]{@{}l@{}}Total \\ Hours\end{tabular} \\ \hline
English  & 100          & 4,097                                                  & 70,465                                                 & 74,562                                                  & 18.59                                                 & 250.38                                                & 268.97                                                 \\

German   & 98           & 4,016                                                  & 81,211                                                 & 85,227                                                  & 16.57                                                 & 264.75                                                & 281.32                                                 \\
Italian  & 100          & 4,047                                                  & 84,587                                                 & 88,634                                                  & 15.69                                                 & 252.28                                                & 267.96                                                 \\
French   & 98           & 3,899                                                  & 70,430                                                 & 74,329                                                  & 16.30                                                 & 234.92                                                & 251.22                                                 \\
Spanish  & 100          & 4,058                                                  & 72,889                                                 & 76,947                                                  & 16.76                                                 & 237.63                                                & 254.38                                                 \\

 \hline
\textbf{Total} & 496          & 20,117                                                 & 379,582                                                & 399,699                                                 & 83.90                                                 & 1,239.95                                               & 1,323.85                                                \\ \hline
\end{tabular}
}
\caption{\ourname composition by language.}
\label{tab:language-distribution}
\end{table}

\begin{figure*}[t]
    \centering
    \includegraphics[width=0.9\linewidth]{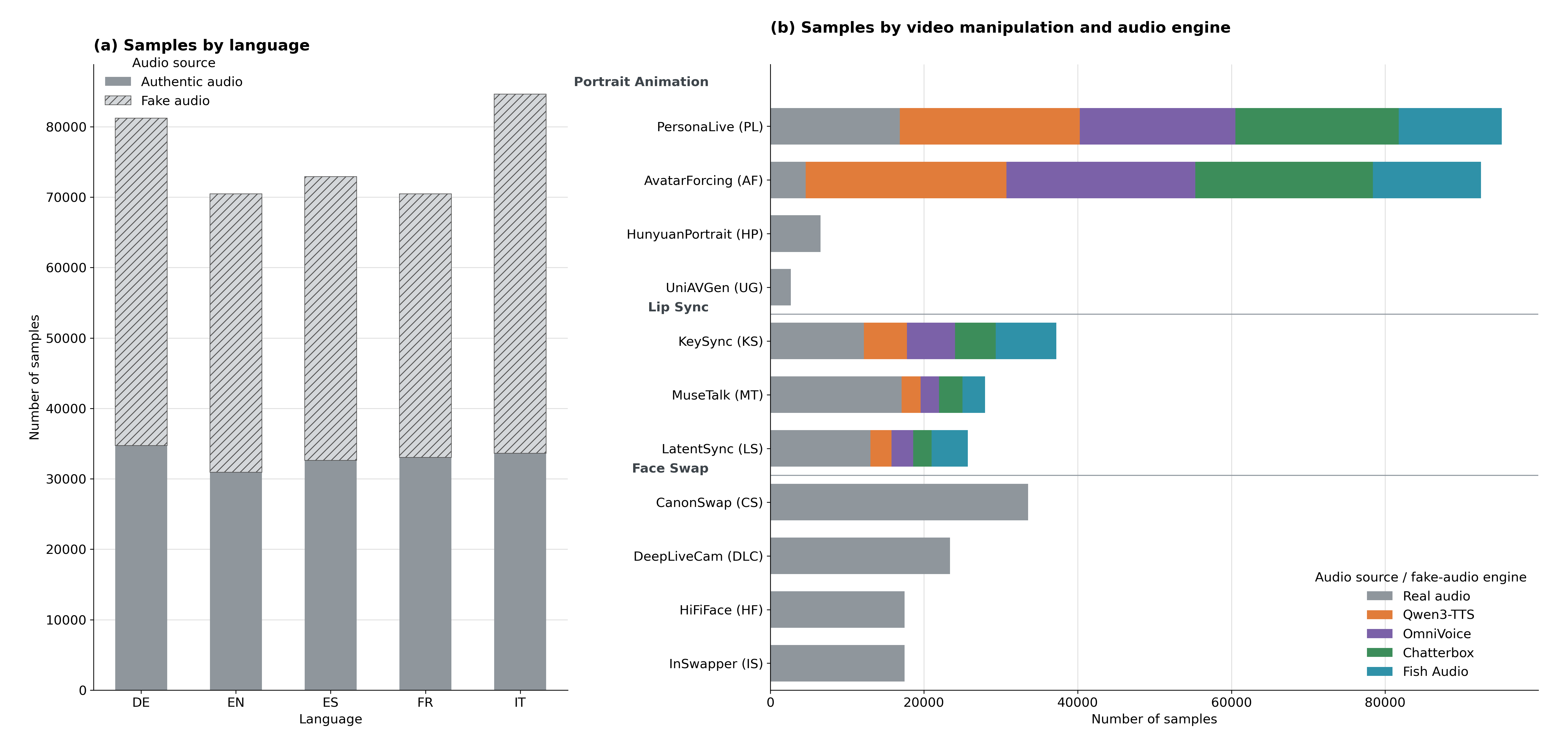}
    \vspace{-5mm}
    \caption{Composition of the generated \ourname benchmark.
    (a) Distribution of fake samples across the five supported languages. Gray bars correspond to videos with authentic audio, while hatched bars denote videos with synthesized speech.
    (b) Distribution of generated samples across the three video manipulation categories. Colors indicate the speech synthesis model paired with each manipulation method (PL~\cite{personalive}, AF~\cite{avatarforcing}, HP~\cite{hunyuanportrait}, UG~\cite{uniavgen}, KS~\cite{keysync}, MT~\cite{musetalk}, LS~\cite{latentsync}, CS~\cite{canonswap}, DLC~\cite{deeplivecam}, HF~\cite{HifiFace}, IS~\cite{inswapper}). Face-swapping methods preserve the authentic audio, whereas portrait animation and lip synchronization are available with both authentic and synthesized speech.}
    \label{fig:samples_breakdown}    
\end{figure*}

To better understand the proposed benchmark, we analyze it from multiple perspectives, including language distribution, clip duration, and manipulation diversity. Overall, the dataset contains $\sim$399k video clips, subdivided into $\sim$20k real and $\sim$379k fake videos, spanning $5$ languages and $3$ classes of visual plus audio manipulations. 

\subsection{Demographic Diversity}
The demographic diversity of a~\DF\ benchmark is an important factor for evaluating model robustness and reducing population-specific biases. Beyond balancing the number of speakers across languages,~\ourname includes participants with diverse demographic backgrounds in terms of age, sex, ethnicity, and geographic origin. Participants range from 18 to 78 years old, with a median age of 32, and exhibit a nearly balanced sex distribution (49.6\% female and 50.4\% male). The dataset also includes participants from different ethnic backgrounds, with 72.2\% identifying as White, 11.1\% Mixed, 8.7\% Black, 3.6\% Asian, and 4.0\% other ethnicities.

~\Cref{fig:nationality_top10} presents the geographic distribution of participants across (and within) the $5$ languages. Rather than limiting each language to a single region, the dataset includes speakers from diverse regions worldwide. For example, the English bucket includes speakers from the United Kingdom, the United States, Canada, South Africa, and several other countries, while the Spanish and French subsets similarly comprise participants from multiple national backgrounds. This diversity reduces the correlation between language and nationality, encouraging models to prioritize the manipulation artifacts over demographic or geographic cues.

Overall, ~\ourname includes participants from 45 countries, providing substantially broader geographic coverage than datasets collected from a single linguistic region. 

\subsection{Language and Manipulation Distribution} %
As summarized in Tab.~\ref{tab:language-distribution} and Fig.~\ref{fig:samples_breakdown},~\ourname is balanced across $5$ supported languages, with $98–100$ unique participants per language; the real and manipulated samples are roughly balanced per language. This composition minimizes language-specific bias and supports multilingual and cross-lingual evaluation.
Fig.~\ref{fig:samples_breakdown}(b) illustrates the distribution across manipulations. The benchmark combines three complementary video manipulation categories, portrait animation, lip synchronization, and face swapping, covering $11$ state-of-the-art generation methods. Portrait animation and lip synchronization are available with both authentic and synthesized speech, whereas face swapping preserves the original audio, enabling controlled comparisons between visual-only and audio-visual manipulations.

\subsection{Clip Duration}
\begin{figure}[t]
    \centering
    \includegraphics[width=\linewidth]{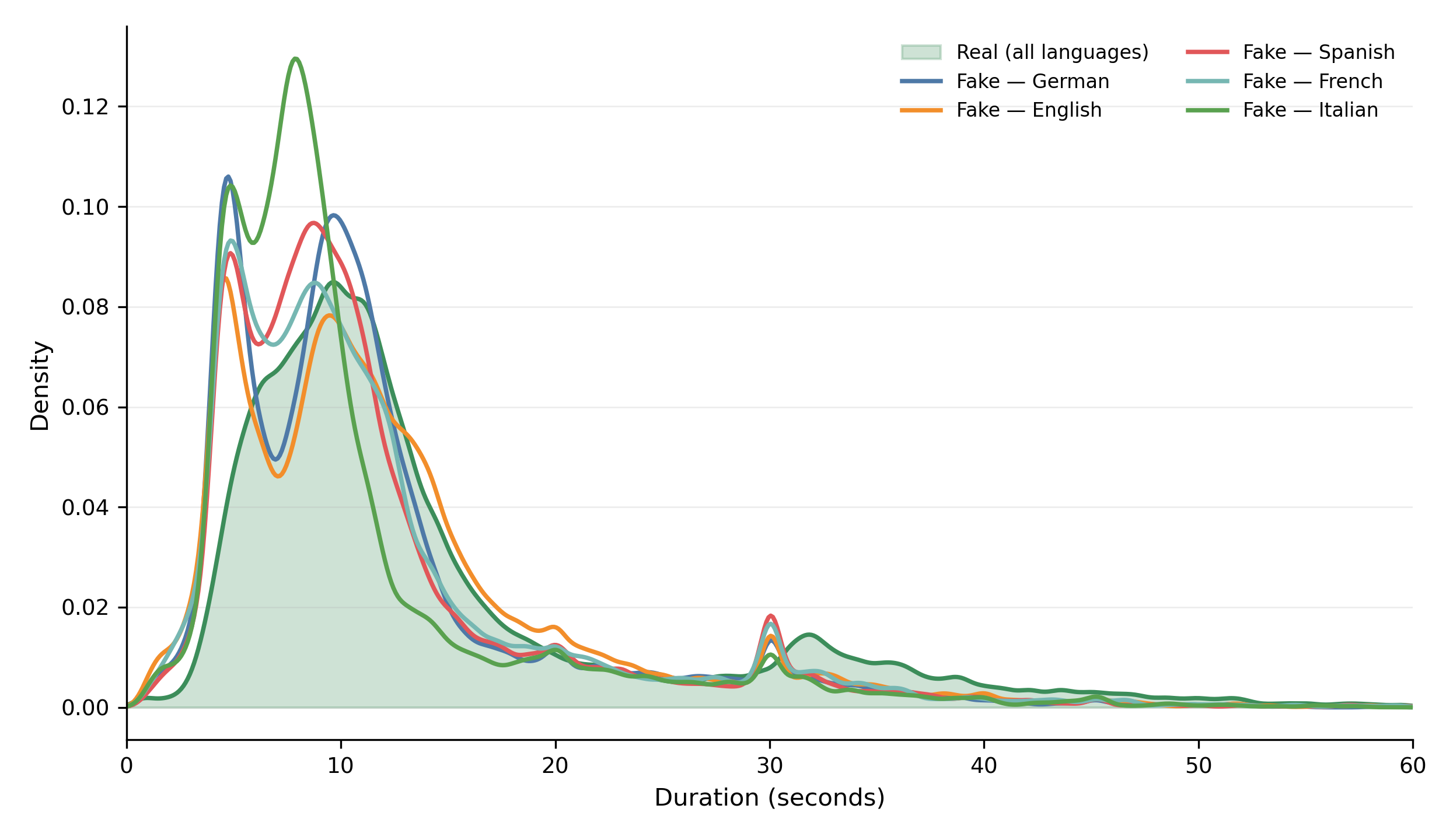}
    \vspace{-6mm}
    \caption{Distribution of clip durations in \ourname. The shaded area represents the real recordings; the colored curves correspond to manipulated videos (by language). The similarity between the real and fake distributions shows that the DeepFakes largely preserve the temporal characteristics of the source recordings.}
    \label{fig:clip_duration}
\end{figure}

Fig.~\ref{fig:clip_duration} shows the distribution of clip durations in \ourname. The shaded regions correspond to the set of real recordings, encompassing all languages, while the colored curves show the duration distributions of the manipulated samples, divided by language. Most clips are between 5 and 15 seconds long, reflecting the combination of standardized reading tasks and short spontaneous responses in the recording protocol. A small group of clips is around 30 seconds, and corresponds to the longer open-ended tasks. 

\section{Benchmark Evaluation}

\begin{table*}[ht]
\resizebox{\linewidth}{!}{%
\begin{tabular}{lll|ll|ll|l|l|ll|ll|ll|l|l|l|l}
\textbf{Model}                                & \textbf{\begin{tabular}[c]{@{}c@{}}Training\\ Data\end{tabular}} & \textbf{AUC}         & \multicolumn{2}{l|}{PersonaLive}                                                                    & \multicolumn{2}{l|}{AvatarForcing}                                                                  & \begin{tabular}[c]{@{}l@{}}Hunyuan\\ Portrait\end{tabular} & \begin{tabular}[c]{@{}l@{}}UniAV\\ Gen\end{tabular} & \multicolumn{2}{l|}{KeySync}                                                                        & \multicolumn{2}{l|}{MuseTalk}                                                                        & \multicolumn{2}{l|}{LatentSync}                                                                      & \begin{tabular}[c]{@{}l@{}}Canon\\ Swap\end{tabular} & \begin{tabular}[c]{@{}l@{}}DeepLive\\ Cam\end{tabular} & \begin{tabular}[c]{@{}l@{}}HiFi\\ Face\end{tabular}                                    & \begin{tabular}[c]{@{}l@{}}In\\ Swapper\end{tabular} \\ \hline
                                              &                   &                      & \begin{tabular}[c]{@{}l@{}}Fake \\ Audio\end{tabular} & \begin{tabular}[c]{@{}l@{}}Real \\ Audio\end{tabular} & \begin{tabular}[c]{@{}l@{}}Fake \\ Audio\end{tabular} & \begin{tabular}[c]{@{}l@{}}Real\\ Audio\end{tabular} & \begin{tabular}[c]{@{}l@{}}Real \\ Audio\end{tabular}              & \begin{tabular}[c]{@{}l@{}}Real \\ Audio\end{tabular}        & \begin{tabular}[c]{@{}l@{}}Fake \\ Audio\end{tabular} & \begin{tabular}[c]{@{}l@{}}Real \\ Audio\end{tabular} & \begin{tabular}[c]{@{}l@{}}Fake \\ Audio\end{tabular} & \begin{tabular}[c]{@{}l@{}}Real \\ Audio\end{tabular} & \begin{tabular}[c]{@{}l@{}}Fake \\ Audio\end{tabular} & \begin{tabular}[c]{@{}l@{}}Real \\ Audio\end{tabular} & \begin{tabular}[c]{@{}l@{}}Real \\ Audio\end{tabular}         & \begin{tabular}[c]{@{}l@{}}Real \\ Audio\end{tabular}           & \begin{tabular}[c]{@{}l@{}}Real \\ Audio\end{tabular} & \begin{tabular}[c]{@{}l@{}}Real \\ Audio\end{tabular}         \\ \hline
\multicolumn{1}{l|}{\bn-A}        & \multicolumn{1}{l|}{DeepSpeak}         & 60.09                & 68.49                                                 & 51.70                                                 & 67.22                                                 & 49.79                                                & 53.58                                                              & 38.05                                                        & \textbf{68.75}                                        & 50.51                                                 & 64.52                                                 & 50.62                                                 & 69.45                                                 & 50.70                                                  & 49.79                                                         & 49.46                                                           & 49.79                                                 & 49.75                                                         \\
\multicolumn{1}{l|}{\bn-V}        & \multicolumn{1}{l|}{DeepSpeak}         & 70.31                & 67.16                                                 & 67.96                                                 & 76.83                                                 & 63.73                                                & 66.51                                                              & 54.10                                                        & 64.97                                                 & 54.57                                                 & \textbf{87.79}                                        & \textbf{81.44}                                        & 75.33                                                 & 59.92                                                 & 72.07                                                         & 67.92                                                           & 60.45                                                 & 67.11                                                         \\
\multicolumn{1}{l|}{\bn-AV}       & \multicolumn{1}{l|}{DeepSpeak}         & {\ul 72.05}          & 81.65                                                 & 72.11                                                 & 75.41                                                 & 61.51                                                & 66.84                                                              & 52.13                                                        & 66.27                                                 & 53.42                                                 & 83.18                                                 & 73.87                                                 & 76.26                                                 & 57.57                                                 & 69.49                                                         & 65.96                                                           & 58.92                                                 & 65.81                                                         \\ \hline
\multicolumn{1}{l|}{\bn-A}        & \multicolumn{1}{l|}{MAVOS-DD}           & 43.93                & 38.14                                                 & 48.32                                                 & 39.69                                                 & 51.89                                                & 48.97                                                              & 65.72                                                        & 41.54                                                 & 46.10                                                 & 44.21                                                 & 47.57                                                 & 42.74                                                 & 47.96                                                 & 51.38                                                         & 50.65                                                           & 48.78                                                 & 48.72                                                         \\
\multicolumn{1}{l|}{\bn-V}        & \multicolumn{1}{l|}{MAVOS-DD}          & 72.56                & 71.93                                                 & 72.06                                                 & 75.11                                                 & 73.03                                                & 66.88                                                              & 69.90                                                        & 59.89                                                 & 56.44                                                 & 77.90                                                 & 77.41                                                 & 74.15                                                 & \textbf{60.75}                                                 & 79.92                                                         & 76.60                                                           & 69.82                                                 & 74.67                                                         \\
\multicolumn{1}{l|}{\bn-AV}       & \multicolumn{1}{l|}{MAVOS-DD}         & {\ul 73.88}          & 73.16                                                 & 72.65                                                 & 77.07                                                 & 74.62                                                & 68.90                                                              & 68.51                                                        & 62.47                                                 & 56.85                                                 & 80.81                                                 & 78.91                                                 & 75.57                                                 & 60.49                                                 & \textbf{81.51}                                                & \textbf{77.30}                                                  & 69.57                                                 & 75.05                                                \\ \hline
\multicolumn{1}{l|}{AVH-Align}       & \multicolumn{1}{l|}{AV1M}               & 18.51                & 33.71                                                 & 19.07                                                 & 4.83                                                  & 5.04                                                 & 35.56                                                              & 28.84                                                        & 9.05                                                  & 4.56                                                  & 29.24                                                 & 6.52                                                  & 12.63                                                 & 6.56                                                  & 36.03                                                         & 39.42                                                           & 9.88                                                  & 9.29                                                          \\ \hline
\multicolumn{1}{l|}{AVH-Align Sup}   & \multicolumn{1}{l|}{FakeAVCeleb}       & {\ul 71.37}          & 70.80                                                 & 82.68                                                 & 77.49                                                 & \textbf{84.24}                                       & 70.81                                                              & 47.27                                                        & 53.28                                                 & \textbf{57.75}                                        & 65.80                                                 & 78.11                                                 & 58.43                                                 & 57.65                                                 & 75.13                                                         & 73.77                                                           & 70.11                                                 & 77.85                                                         \\
\multicolumn{1}{l|}{AVH-Align Sup}   & \multicolumn{1}{l|}{MAVOS-DD}          & 67.11                & 75.48                      & 65.42                          & 69.19                                                 & 55.36                                                & \textbf{73.15}                                                              & 62.23                                                        & 50.89                                                 & 36.08                                                 & 75.73                                                 & 59.95                                                 & 75.00                                                 & 43.82                                                 & 78.60                                                         & 71.80                                                           & 57.31                                                 & 66.16                                                         \\
\multicolumn{1}{l|}{AVH-Align Sup}   & \multicolumn{1}{l|}{AV1M}              & 41.26                & 35.33                                                 & 51.97                                                 & 30.91                                                 & 27.92                                                & 50.16                                                              & 40.41                                                        & 35.50                                                 & 25.43                                                 & 38.02                                                 & 29.86                                                 & 39.75                                                 & 34.43                                                 & 49.96                                                         & 50.35                                                           & \textbf{77.95}                                        & \textbf{78.34}                                                         \\ \hline
\multicolumn{1}{l|}{SpeechForensics} & \begin{tabular}[c]{@{}l@{}}VoxCeleb2\\ \& LRS3\end{tabular}          & \multicolumn{1}{|l|}{{\ul\textbf{79.79}}} & \textbf{97.01}                                        & \textbf{87.71}                                        & \textbf{85.45}                                        & 75.71                                                & 69.68                                                     & \textbf{78.22}                                               & 67.10                                                 & 43.75                                                 & 84.86                                                 & 69.86                                                 & \textbf{82.30}                                         & 49.29                                        & 80.22                                                         & 72.71                                                           & 63.62                                                 & 70.98                                                         \\ \hline
\multicolumn{1}{l|}{\textit{Avg. Audio}}                                    & -                 & \multicolumn{1}{|l|}{52.01}                                                             & 53.31                                                & 50.01  & 53.46 & 50.84 &51.28 &51.89 &55.15 & 48.31 & 54.36 & 49.10 & 56.10 & 49.33 & 50.59 & 50.06   & 49.29 & 49.24                                                                                                    \\
\multicolumn{1}{l|}{\textit{Avg. Video}}                                    & -                 & \multicolumn{1}{|l|}{71.44}                                                             & 69.54                                                 & 70.01  & 75.97                                                & 68.38                                              & 66.70                                                              & 62.00                                                        & 62.43                                                 & 55.51                                                 & 82.85                                                 & 79.43                                                 & 74.74                                                 & 60.34                                                 & 76.00  & 72.26                                                        & 65.14                                                           & 70.89                                                                                                         \\
\multicolumn{1}{l|}{\textit{Avg. AV}}                                       & -                & \multicolumn{1}{|l|}{74.27} & 80.65                                                & 78.79                  & 78.85                                                 & 74.02                                                                                                & 69.06                                                              & 61.53                                                        & 62.28                                                 & 52.94                                                 & 78.66                                                 & 75.19                                                 & 73.14                                                 & 58.57                                                 & 76.59                                                         & 72.44                                                           & 65.56                                                 & 72.42                                                        
\end{tabular}
}
\vspace{-2mm}
\caption{Performance of different \DF\ detectors on the \ourname dataset. Models (\bn~\cite{braven}, AVH-Align~\cite{AVH-Align}, SpeechForensics~\cite{speechforensics}) are evaluated using AUC on the test set, both overall and for each manipulation method (in order, PL~\cite{personalive}, AF~\cite{avatarforcing}, HP~\cite{hunyuanportrait}, UG~\cite{uniavgen}, KS~\cite{keysync}, MT~\cite{musetalk}, LS~\cite{latentsync}, CS~\cite{canonswap}, DLC~\cite{deeplivecam}, HF~\cite{HifiFace}, IS~\cite{inswapper}). Bold values indicate the best AUC for each manipulation. \textit{Avg. Audio} and \textit{Avg. Video} report the average AUC across all audio-only and video-only models, respectively, while \textit{Avg. AV} reports the average AUC across the best-performing audio-visual variants of each detector (underscored). }
\label{tab:testset_results}
\end{table*}

\subsection{Experimental Setup}

We partition \ourname into identity-disjoint training, validation, and test sets with a 70/10/20 split. We set up the evaluation using three different state-of-the-art models.

\textbf{SpeechForensics}~\cite{speechforensics} is an unsupervised approach that leverages synchronization to detect DeepFakes. It extracts AV-HuBERT~\cite{avhubert} audio-visual features and evaluates their temporal consistency. The model is trained on LRS3~\cite{lrs3dataset} and VoxCeleb2~\cite{voxceleb} and, due to its unsupervised nature, does not require any DeepFake-specific fine-tuning. Similarly, \textbf{AVH-Align}~\cite{AVH-Align} leverages AV-HuBERT representations and is trained on real videos to maximize audio-visual embedding alignment. It also supports a probing layer for supervised \DF\ detection. For the unsupervised baseline, we use the public checkpoints trained on AV-DeepFake-1M~\cite{av1m} (AV1M). For the supervised setting, we train AVH-Align on MAVOS-DD~\cite{mavos} and use the available checkpoints on FakeAVCeleb~\cite{FakeAVCeleb} and AV-DeepFake-1M. Finally, \textbf{(B)RAVEn}~\cite{braven} is a self-supervised representation-learning framework, adapted for \DF\ detection via linear probing. We keep the audio-visual encoder frozen and train a single linear layer for binary real/fake classification, considering audio-only (-A), video-only (-V), and audio-visual (-AV) variants. The classification heads are trained on both DeepSpeak~\cite{DeepSpeak} and MAVOS-DD. 
Further details are presented in~\cref{app:exp_setup}.

\subsection{Evaluation Results on \ourname}
~\cref{tab:babel_training} reports in-domain results, showing that it is ``solvable'' when training/testing in-domain. More importantly,  ~\Cref{tab:testset_results} reports the cross-dataset evaluation of state-of-the-art detectors on the test set of the \ourname benchmark. 

\textbf{Overall Performance.} SpeechForensics achieves the best overall AUC (79.79), showing robust generalization. It performs well on fully synthetic inputs, peaking on PersonaLive~\cite{personalive} (97.01) and AvatarForcing~\cite{avatarforcing} (85.45) with fake audio. However, performance degrades significantly on manipulations paired with \emph{real} audio, such as KeySync~\cite{keysync} (43.75) and HiFiFace~\cite{HifiFace} (63.62), showing the difficulty of detecting visual artifacts when audio cues remain pristine. %

\textbf{Multimodal vs Unimodal.} Multimodal evaluation consistently yields more balanced detection performance. Under DeepSpeak training, \bn-AV achieves 72.05 AUC, outperforming its audio-only (60.09) and video-only (70.31) versions. Similarly, with MAVOS-DD, \bn-AV obtains the highest score among the probed models. This trend is consistent across the macro-averages, where multimodal architectures achieve an AUC of 74.27, compared to 71.44 for video-only, confirming that complementary cross-modal information enhances generalization.

\textbf{Impact of Training Data.} Generalization is highly sensitive to the training dataset. For \bn, MAVOS-DD improves the video-only and audio-visual variants over DeepSpeak, while the audio-only model transfers substantially better from DeepSpeak. Training on AV-DeepFake-1M seems to fail, potentially due to its focus on partially manipulated samples rather than fully fake ones. At the same time, AVH-Align Supervised on FakeAVCeleb generalizes effectively (71.37 AUC), achieving the highest performance on the AvatarForcing subset with real audio, which is challenging for the other models. 

\textbf{Weakness on Real-Audio Manipulations.} Performance varies drastically across generation techniques. In particular, KeySync~\cite{keysync} and UniAVGen~\cite{uniavgen} are harder to detect, on average. Crucially, a consistent degradation emerges across almost every evaluated method: detection accuracy drops in the \emph{real-audio} scenario compared to the fake-audio. This reveals a common weakness of current detectors when the original audio is preserved. Interestingly, this degradation also affects video-only models, suggesting that the resulting video quality is impacted by using fake audio during synthesis. Overall, video manipulations with real-audio constitute a particularly challenging setting for current detection approaches. 

\subsection{Multilingual Generalization Results}
\begin{table}[t]
\centering
\resizebox{\linewidth}{!}{%
\begin{tabular}{ll|ccccc}
\textbf{Model} & \textbf{Training data} & \textbf{EN} & \textbf{DE} & \textbf{IT} & \textbf{FR} & \textbf{ES} \\ \hline
\bn-A      & DeepSpeak & 56.79 & 62.46 & 60.66 & 59.83 & 64.41 \\
\bn-V      & DeepSpeak & 74.84 & 71.38 & 69.99 & 68.16 & 70.22 \\
\bn-AV     & DeepSpeak & 78.73 & 74.73 & 71.37 & 70.35 & 73.49 \\ \hline
\bn-A      & MAVOS-DD      & 43.30 & 45.22 & 44.58 & 45.46 & 41.69 \\
\bn-V      & MAVOS-DD      & 71.50 & 70.63 & 75.83 & 74.64 & 71.09 \\
\bn-AV     & MAVOS-DD      & 73.09 & 72.38 & 77.41 & 75.57 & 72.05 \\ \hline
AVH-Align     & AV1M          & 21.46 & 17.28 & 18.07 & 16.88 & 19.22 \\ \hline
AVH-Align Sup & MAVOS-DD      & 64.04	& 69.05 & 	66.88& 	66.31	& 68.77 \\
AVH-Align Sup & FakeAVCeleb   & 75.55 & 76.37 & 68.40 & 69.47 & 67.09 \\
AVH-Align Sup & AV1M          & 43.94 & 43.03 & 39.44 & 42.34 & 37.46 \\ \hline
SpeechForensics & VoxCeleb2\&LRS3     & \textbf{80.99} & \textbf{78.91} & \textbf{77.54} & \textbf{77.73} & \textbf{83.83} \\ 
\end{tabular}
}
\vspace{-2mm}
\caption{Per-language AUC on the \ourname benchmark. Results are reported separately for English (EN), German (DE), Italian (IT), French (FR), and Spanish (ES). Bold values indicate the best performance for each language.}
\label{tab:polyspeak_language}
\end{table}

~\cref{tab:polyspeak_language} reports per-language performance across English (EN), German (DE), Italian (IT), French (FR), and Spanish (ES). Individual models exhibit different performance patterns across languages, with no language consistently yielding the highest or lowest AUC. This highlights the role of \ourname as a multilingual benchmark for exposing language-dependent variations in \DF\ detectors.

\textbf{Overall Performance.} SpeechForensics achieves the highest AUC across all five languages, peaking at 83.83 on Spanish and reaching 80.99 on English. The model demonstrates consistent cross-lingual generalization, maintaining an AUC above 77.50 across all language splits. Furthermore, across both \bn training schemes, multimodal modeling (-AV) consistently improves over video-only representations (-V) in every language evaluated. Correlation analysis shows that performance on EN diverges  from the other languages, highlighting its ``special'' status in the training data. EN correlates the most with DE, another Germanic language. At the same time, the Romance languages (IT, FR, ES) show strong mutual correlation.

\textbf{Training Data Bias.} Training distributions influence linguistic transfer differently across detector architectures. For \bn, MAVOS-DD training yields stronger multimodal transfer on Romance languages such as Italian (77.41 vs. 71.37) and French (75.57 vs. 70.35), whereas DeepSpeak yields higher transfer on English (78.73 vs. 73.09) and German (74.73 vs. 72.38). For AVH-Align, FakeAVCeleb supervision provides the strongest representations overall, consistently outperforming MAVOS-DD and AV-DeepFake-1M on almost all languages. Across all splits, AV-DeepFake-1M yields the lowest transferability, likely due to being focused on local manipulations. %

\textbf{Language Robustness.} No individual language proves systematically harder or easier across detectors. \bn trained on MAVOS-DD performs the best on Italian, while SpeechForensics peaks on Spanish, and AVH-Align supervised on FakeAVCeleb performs the best on German. The absence of a consistent ranking across languages suggests that language sensitivity depends on the detector architecture and training data, highlighting the importance of multilingual evaluation for assessing detector robustness.

\subsection{Performance Across Demographic Subgroups}

\begin{figure}
    \centering
    \includegraphics[width=\linewidth]{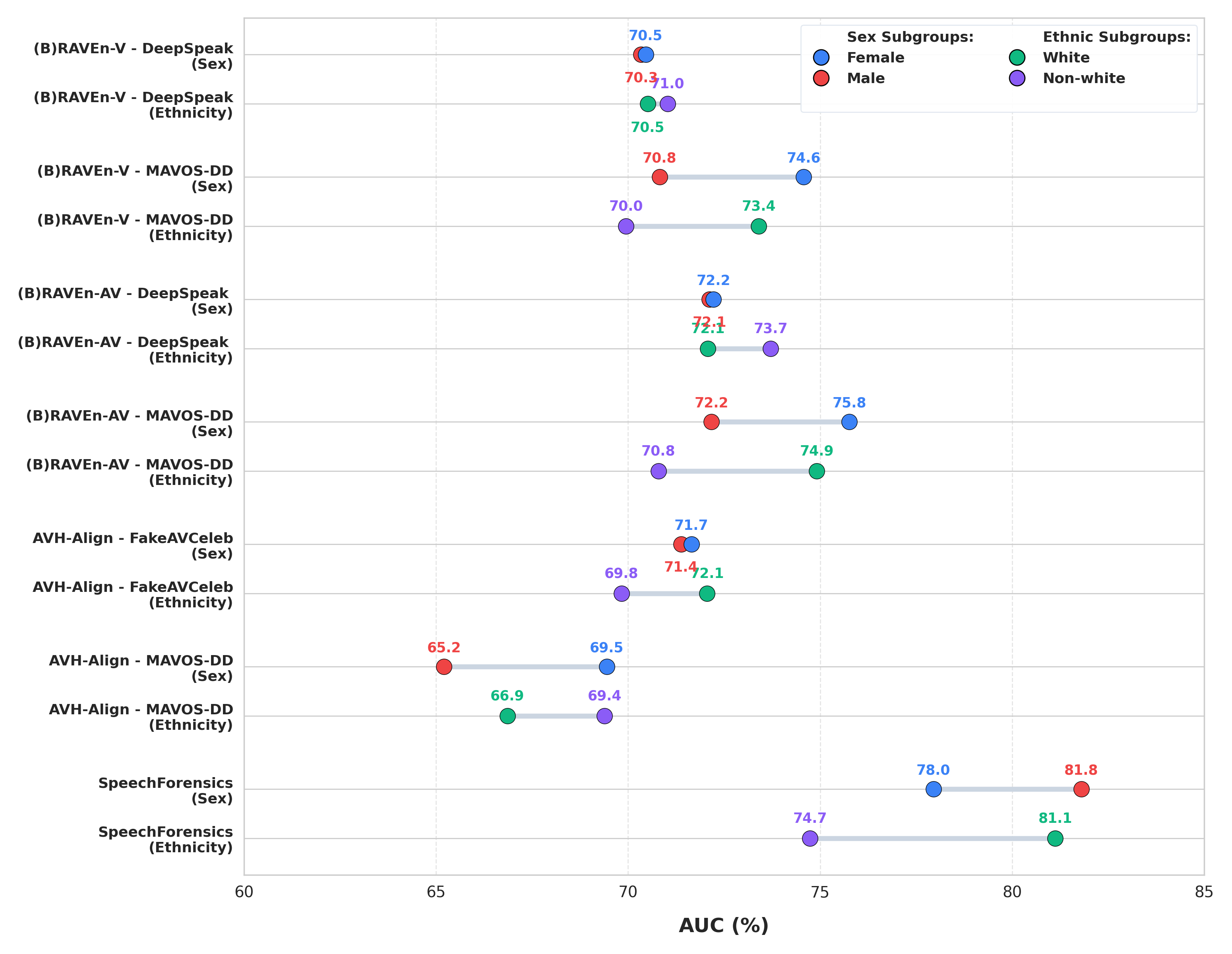}
    \vspace{-6mm}
    \caption{Demographic bias analysis across sex (Female vs. Male) and ethnicity (White vs. Non-white) subgroups on the \ourname test set. Points report the AUC per subgroup, with connecting bars indicating the performance gap across detector backbones and training datasets.}
    \label{fig:bias_analysis}
\end{figure}

\ourname's design allows us to study whether detection methods exhibit systematic demographic biases. For that, we evaluate detector performance across reported sex (Female, Male) and ethnicity (White, Non-white) partitions. Fig.~\ref{fig:bias_analysis} shows the subgroup AUC distributions for each detector. In most cases, performance remains fairly balanced across partitions. Most subgroup disparities however remain below 4 AUC points.%

\textbf{Sex and Ethnicity Disparities.} Gaps in sex are generally tighter than in ethnicity. For instance, \bn-AV trained on DeepSpeak exhibits nearly identical accuracy across female and male subjects, and AVH-Align supervised on FakeAVCeleb shows a minor $0.3$ AUC gap. Larger differences appear under MAVOS-DD training, where both \bn-AV (75.8 vs 72.2) and AVH-Align (69.5 vs 65.2) perform significantly better on female subjects, with gaps of $3.6$ and $4.3$ AUC points, respectively. Similar behavior emerges across ethnicity partitions: SpeechForensics exhibits the largest disparity across the benchmark, achieving 81.1 AUC on White subjects compared to 74.7 on Non-white subjects ($\Delta=6.4$).

\textbf{Impact of Training.}  Demographic disparities also vary substantially with the training distribution. Models trained on DeepSpeak exhibit small subgroup gaps, remaining below $1.6$ AUC points across both sex and ethnicity. In contrast, MAVOS-DD training produces larger disparities for both unimodal and multimodal \bn variants, consistently favoring Female and White subjects.

\subsection{Human Evaluation}

\begin{table}[t]
\small
\centering
\resizebox{\linewidth}{!}{%
\begin{tabular}{llcccc}
\hline
\multicolumn{2}{l}{\textbf{Engine}} & \textbf{\begin{tabular}[c]{@{}c@{}} \# Rated\\ Samples\end{tabular}}          & \textbf{Mean Score}      & \textbf{\begin{tabular}[c]{@{}c@{}}Deception Rate\\ (\%)\end{tabular}} & {\color[HTML]{F56B00} \textbf{\begin{tabular}[c]{@{}c@{}}AUC\\ (\%)\end{tabular}}} \\ \hline
\multicolumn{2}{l}{Real (Baseline)} & 210                     & 1.59                    & 87.60                                                                   & {\color[HTML]{F56B00} -}                                                           \\ \hline
\multicolumn{2}{l}{InSwapper}       & 20                      & 2.45                    & 50.0                                                                   & {\color[HTML]{F56B00} 72.42}                                                       \\\hline
\multicolumn{2}{l}{CanonSwap}       & 18                      & 2.50                    & 50.0                                                                   & {\color[HTML]{F56B00} 76.59}                                                       \\\hline
\multicolumn{2}{l}{DeepLiveCam}     & 19                      & 2.58                    & 42.10                                                                  & {\color[HTML]{F56B00} 72.44}                                                       \\\hline
\multicolumn{2}{l}{AvatarForcing}   & 141                     & 2.62                    & 47.52                                                                   & {\color[HTML]{F56B00} -}                                                           \\
\multicolumn{2}{r}{Real Audio}      & \multicolumn{1}{c}{23}  & \multicolumn{1}{c}{2.57} & \multicolumn{1}{c}{43.48}                                              & \multicolumn{1}{l}{{\color[HTML]{F56B00} 74.02}}                                   \\
\multicolumn{2}{r}{Fake Audio}       & \multicolumn{1}{c}{118} & \multicolumn{1}{c}{2.63} & \multicolumn{1}{c}{48.31}                                              & \multicolumn{1}{l}{{\color[HTML]{F56B00} 78.85}}                                   \\ \hline
\multicolumn{2}{l}{KeySync}         & 143                     & 2.72                    & 43.40                                                                  & {\color[HTML]{F56B00} -}                                                           \\
\multicolumn{2}{r}{Real Audio}      & \multicolumn{1}{c}{30}  & \multicolumn{1}{c}{2.37} & \multicolumn{1}{c}{63.33}                                              & \multicolumn{1}{l}{{\color[HTML]{F56B00} 52.94}}                                   \\
\multicolumn{2}{r}{Fake Audio}      & \multicolumn{1}{c}{113} & \multicolumn{1}{c}{2.81} & \multicolumn{1}{c}{38.05}                                              & \multicolumn{1}{l}{{\color[HTML]{F56B00} 62.28}}                                   \\\hline
\multicolumn{2}{l}{HunyuanPortrait}     & 20                      & 2.71                    & 43.48                                                                   & {\color[HTML]{F56B00} 69.06}  \\ \hline       

\multicolumn{2}{l}{MuseTalk}        & 121                     & 2.73                    & 43.80                                                                  & {\color[HTML]{F56B00} -}                                                           \\
\multicolumn{2}{r}{Real Audio}      & \multicolumn{1}{c}{30}  & \multicolumn{1}{c}{2.67} & 50.00                                                                  & \multicolumn{1}{l}{{\color[HTML]{F56B00} 75.19}}                                   \\
\multicolumn{2}{r}{Fake Audio}      & \multicolumn{1}{c}{87}  & \multicolumn{1}{c}{2.75} & 41.38                                                                  & \multicolumn{1}{l}{{\color[HTML]{F56B00} 78.66}}                                   \\\hline
\multicolumn{2}{l}{LatentSync}      & 108                     & 2.80                    & 38.00                                                                     & {\color[HTML]{F56B00} -}                                                           \\
\multicolumn{2}{r}{Real Audio}      & \multicolumn{1}{c}{30}  & \multicolumn{1}{c}{3.17} & 66.67                                                                  & \multicolumn{1}{l}{{\color[HTML]{F56B00} 58.57}}                                   \\
\multicolumn{2}{r}{Fake Audio}      & \multicolumn{1}{c}{78}  & \multicolumn{1}{c}{3.05} & 26.92                                                                  & \multicolumn{1}{l}{{\color[HTML]{F56B00} 73.14}}                                   \\\hline
\multicolumn{2}{l}{HiFiFace}        & 27                      & 2.85                    & 37.00                                                                   & {\color[HTML]{F56B00} 65.56}                                                       \\\hline
\multicolumn{2}{l}{UniAVGen}        & 72                      & 3.03                    & 26.40                                                                   & {\color[HTML]{F56B00} 61.53}                                                       \\\hline
\multicolumn{2}{l}{PersonaLive}     & 137                     & 3.48                    & 10.90                                                                   & {\color[HTML]{F56B00} -}                                                           \\
\multicolumn{2}{r}{Real Audio}      & \multicolumn{1}{c}{26}  & \multicolumn{1}{c}{3.07} & 26.92                                                                  & \multicolumn{1}{l}{{\color[HTML]{F56B00} 78.79}}                                   \\
\multicolumn{2}{r}{Fake Audio}       & \multicolumn{1}{c}{111}  & \multicolumn{1}{c}{3.57} & 7.21                                                                   & \multicolumn{1}{l}{{\color[HTML]{F56B00} 80.65}}                                  
\end{tabular}
}
\caption{Human evaluation results on a subset of 1,036 videos from the \ourname benchmark, annotated by 51 participants. Ratings are collected using a four-point scale (1: Real, 2: Probably Real, 3: Probably Fake, 4: Fake). The \emph{Deception Rate} denotes the percentage of manipulated videos rated as Real or Probably Real (scores $\leq 2$). For real videos, the same quantity indicates the percentage correctly perceived as genuine. The \emph{AUC} column reports the average AUC of the multimodal detectors (Avg. AV in Tab.~\ref{tab:testset_results}).}
\label{tab:human_eval}
\end{table}

To complement the detector-based evaluation, we conduct an evaluation with \emph{fluent language speakers} to assess the perceptual realism of the manipulated videos in \ourname, Tab~\ref{tab:human_eval}. The ratings are collected using a $4$-point scale (1: Real, 2: Probably Real, 3: Probably Fake, 4: Fake). Based on these ratings, we define the \emph{deception rate} as the percentage of fake videos perceived as Real or Probably Real (scores $\leq 2$). InSwapper and CanonSwap achieve the highest deception rate (50\%), followed by AvatarForcing (47.5\%), MuseTalk (43.8\%), KeySync (43.4\%), and LatentSync (38.0\%), while UniAVGen (26.40\%) and PersonaLive (10.90\%) are more easily recognized as fake. Participants correctly perceive 87.60\% of real videos as genuine. Preserving authentic audio substantially increases the deception rate for KeySync (38.05\% to 63.33\%), LatentSync (26.92\% to 66.67\%), and PersonaLive (7.21\% to 26.93\%), while having a limited impact on AvatarForcing. Overall, the substantial deception rates across several generation techniques confirm that \ourname contains perceptually convincing manipulations. However, human deception and detector performance show only a moderate inverse correlation ($r=-0.48$), indicating that perceptual realism and machine detection difficulty do not fully align. (e.g., CanonSwap and InSwapper). 
See~\cref{app:human_eval_breakdown} for more results.

\section{Discussion and Limitations}
\label{sec:discussion}

\textbf{Ethical Considerations and Data Distribution.} The creation and dissemination of \DF\ datasets carry inherent risks of misuse. To mitigate these risks while supporting reproducible forensic research, \ourname\ was developed under  institutional ethical guidelines. Unlike datasets that scrape  footage without consent, our participants provided informed consent for their biometric data to be manipulated and distributed for academic purposes. \ourname\ will not be released publicly on the open web; access will be granted behind a strict, non-commercial research license, following institutional verification and a data-use agreement. 

\textbf{Limitations and Future Work.} While \ourname\ significantly broadens the linguistic scope of DeepFake benchmarks by incorporating five languages and participants from 45 countries, it currently focuses on Germanic and Romance languages (English, German, French, Italian, and Spanish). Consequently, the benchmark does not capture the phonetic and visual-speech dynamics of non-Latin-script languages (e.g., Mandarin, Arabic, or Hindi), which remain an important avenue for future extensions. Furthermore, although our pipeline incorporates 15 state-of-the-art generative models, the rapid pace of generative AI means synthesis artifacts will continually evolve.

\section{Conclusion}
\label{sec:conclusion}

In this paper, we introduced \ourname, a consent-first multilingual audio-visual DeepFake benchmark. By combining a controlled webcam acquisition protocol with a modular synthesis pipeline, \ourname\ disentangles visual manipulations from synthetic speech. This enables a systematic diagnostic evaluation of state-of-the-art multimodal detectors. Our benchmarking reveals that detection outcome strongly depends on the audio-visual generation pairing, with substantial performance decrease when authentic audio is preserved. Cross-lingual evaluation shows no language consistently harder across detectors; sensitivity instead varies across architectures and training data. Further, fine-grained demographic evaluation exposes detector- and training-specific subgroup disparities that can be obscured by aggregate performance. Finally, human evaluation shows that perceptual realism and machine-detection difficulty do not fully align. \ourname\ provides a rigorous, ethically sourced foundation for developing fair, language-agnostic, and modality-robust DeepFake detection systems.

\paragraph{Acknowledgments.} We gratefully acknowledge support from the hessian.AI Service Center (funded by the Federal Ministry of Research, Technology and Space, BMFTR, grant no. 16IS22091) and the hessian.AI Innovation Lab (funded by the Hessian Ministry for Digital Strategy and Innovation, grant no. S-DIW04/0013/003).

{
    \small
    \bibliographystyle{ieeenat_fullname}
    \bibliography{main}
}

\clearpage
\appendix
\section{Informed Consent Form}
\label{sec:consent_form}
As the dataset was collected from human participants, all contributors were required to provide informed consent before participating. The consent form describes the purpose of the study, the types of data collected, the intended use of the recordings, the potential risks associated with participation, and participants' rights regarding the deletion of their data. The full consent form used during data collection is presented below.

\begin{tcolorbox}[
enhanced,
breakable,
colback=black!2,
colframe=black!20,
boxrule=0.4pt,
arc=2mm,
left=10pt,
right=10pt,
top=10pt,
bottom=10pt,
borderline west={2pt}{0pt}{black!55},
before skip=10pt,
after skip=10pt
]
Please read the following information carefully.

\subsection*{Purpose of the Study}

This study aims to collect short video responses to create a research dataset for developing and evaluating deepfake detection systems. Deepfakes are videos that have been digitally altered to change a person’s face, voice, expressions, or what they are saying using artificial intelligence.

In this project, we will generate benign deepfake videos purely for academic purposes. These generated videos will not depict misleading, harmful, or inappropriate content.

Your participation will help ensure that future deepfake detection systems are fair, accurate, and effective across different genders, skin tones, languages, and cultural backgrounds.

\subsection*{What You’ll Do}

During the study, you will interact with a chat-like interface that presents a series of prompts. For each prompt, you will record a short video response (including both video and audio) using your camera and microphone.

There are three types of recordings you will be asked to make:

\begin{itemize}
\item \textbf{Reading Aloud:} You will be shown short pieces of text, such as a sentence, a few sentences, or a short paragraph, and asked to record yourself reading them aloud.

\item \textbf{Question Answering:} You will be asked simple, general questions (e.g., about hobbies, daily activities, or opinions on neutral topics) and record your spoken responses.
\item \textbf{Action Prompts:} You will be asked to perform simple gestures (such as nodding, turning your head, or gesturing with your hands) or show a natural facial expression (such as smiling) while speaking a short phrase.

\end{itemize}

All texts, questions, and actions are carefully designed to be neutral and non-offensive. You are not required to share personal or truthful information and may freely invent responses if you prefer.

\subsection*{Use of Public-Domain Texts}

Some reading-aloud prompts are taken from books in the public domain. Whenever a sentence or passage originates from such a source, the title of the book will be displayed alongside the text.

If you encounter a sentence or passage that you believe is inappropriate, outdated, or otherwise problematic, you may flag it within the interface. All flagged items will be reviewed and, if deemed inappropriate, removed from future versions of the study.

\subsection*{Privacy and Data Collection}

We process your data in accordance with the General Data Protection Regulation (GDPR). %

Your video responses will be recorded and stored securely on our local servers. The collected and generated data will only be made available to vetted researchers upon request and will be used exclusively for research purposes.

\paragraph{Data Collected}

\begin{itemize}
\item Video and audio recordings
\item Demographic information (e.g., age, gender, ethnicity, language proficiency)
\item Device information (microphone and webcam characteristics)
\end{itemize}

\subsection*{Risks Associated with Participation}

Although all data will be handled with care, there is a small risk of data misuse, such as unauthorized redistribution or manipulation of the dataset.

To minimize these risks, dataset distribution is governed by a strict license agreement that:

\begin{itemize}
\item Restricts the use of recordings and generated data to academic, non-commercial research.
\item Prohibits redistribution, resale, or repurposing outside approved research contexts.
\item Requires proper attribution and compliance with ethical guidelines.
\end{itemize}

In addition, all data will be stored securely and access will be limited to authorized researchers. While these safeguards substantially reduce the likelihood of misuse, no system can provide an absolute guarantee against unauthorized use once data have been shared.

\subsection*{Your Rights}

Participation in this study is entirely voluntary. You may stop participating at any time or withdraw your consent at a later stage without penalty or disadvantage.

It is important to distinguish between stopping participation before completing the survey and withdrawing consent after completion.

\paragraph{1. If you stop participating during the survey}

\begin{itemize}
\item You may exit the survey at any point before submitting your final responses.
\item Any recordings made up to that point will be permanently deleted.
\item No data from an incomplete session will be used for research purposes.
\end{itemize}

\paragraph{2. If you withdraw your consent after completing the survey}

\begin{itemize}
\item You may request the removal of your data from our storage at any time.
\item Your data will not be included in future releases of the dataset.
\item However, if the dataset has already been distributed to approved researchers, we cannot guarantee that previously distributed copies can be recalled or modified.
\end{itemize}

You have the right to know which data concerning you have been stored and to request correction or deletion where applicable. As noted above, we can only guarantee that future releases of the dataset will reflect deletion or correction requests.

In case of disagreement, you have the right to lodge a complaint with the Data Protection Officer.

\subsection*{Time Commitment}

This survey takes approximately 40 minutes to complete.
\end{tcolorbox}

\subsection{Dataset Access, Governance, and Participant Withdrawal}

\ourname will be distributed through a controlled-access system rather than released publicly. Access will be limited to verified researchers affiliated with academic institutions and granted for research purposes under a data-use agreement. The agreement will prohibit unauthorized redistribution, sharing with non-approved third parties, and misuse of participant identities or recordings.

Access requests and authorized recipients will be recorded to enable communication of dataset updates and participant withdrawal requests. Dataset releases will be versioned to track changes resulting from withdrawals or data corrections.

Participants may request the removal of their data in accordance with the consent procedure. Upon a valid withdrawal request, we will remove the participant's original recordings, associated metadata, and all derived samples that use their face or voice. The participant will consequently be excluded from all future releases of the dataset.

Since distributed copies cannot be recalled, researchers who have already received the dataset will be notified of the withdrawal and, under the data-use agreement, required to delete the corresponding participant IDs and associated data from their local copies. Subsequent dataset versions and newly granted access will exclude the withdrawn participant.

\section{Web Application}
\label{sec:data_coll}

To support large-scale, standardized data collection, we developed a custom web-based application that guides participants through the acquisition process. The platform manages participant registration, informed consent, device verification, and task presentation while recording synchronized audio and video directly within the browser. This design ensures a consistent acquisition protocol across participants, even when recordings are performed remotely on personal devices.

\begin{figure*}[t]
    \centering

    \begin{subfigure}[t]{0.49\textwidth}
        \centering
        \includegraphics[width=\linewidth]{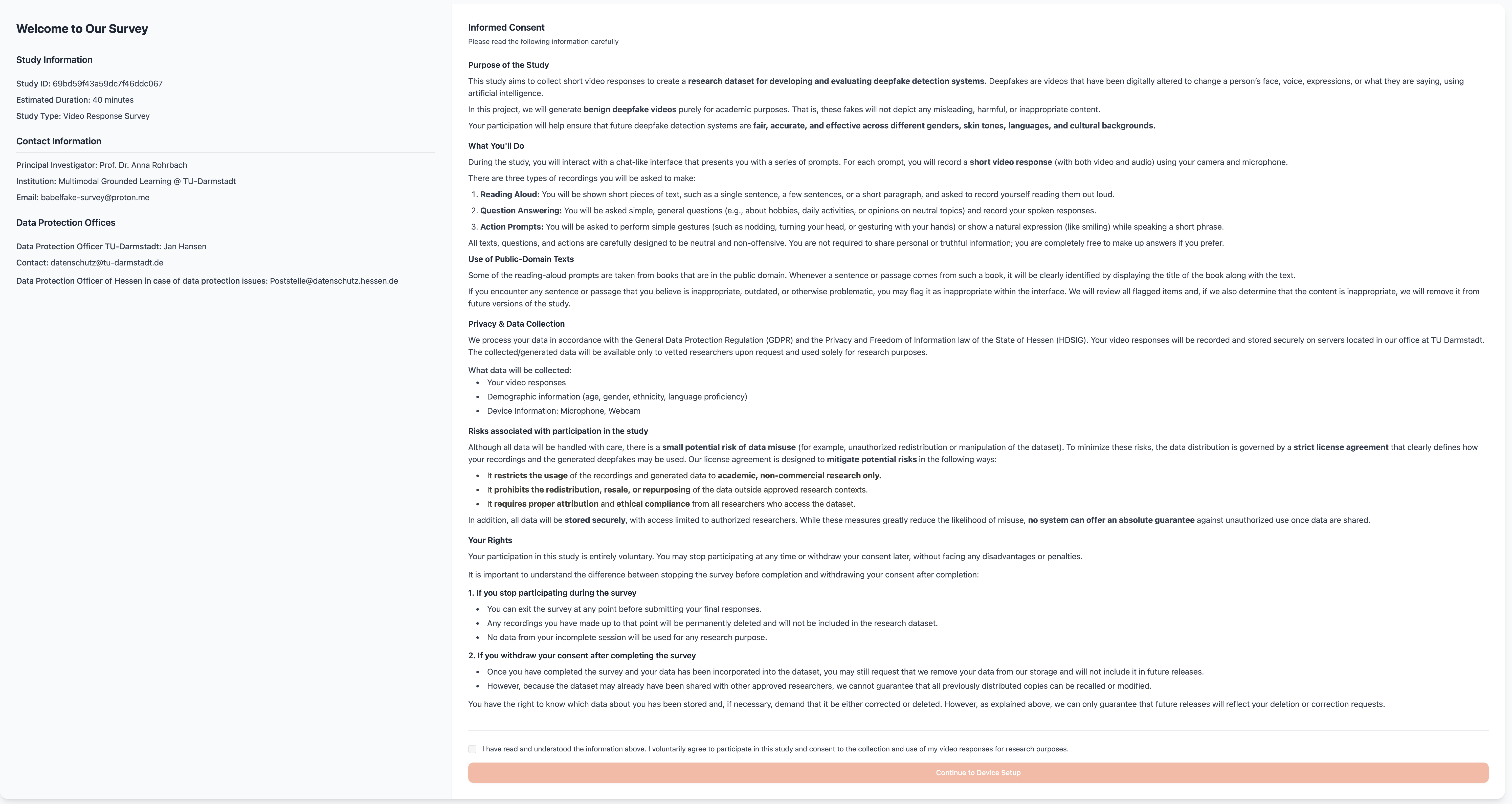}
        \caption{Study information and informed consent.}
    \end{subfigure}
    \hfill
    \begin{subfigure}[t]{0.49\textwidth}
        \centering
        \includegraphics[width=\linewidth]{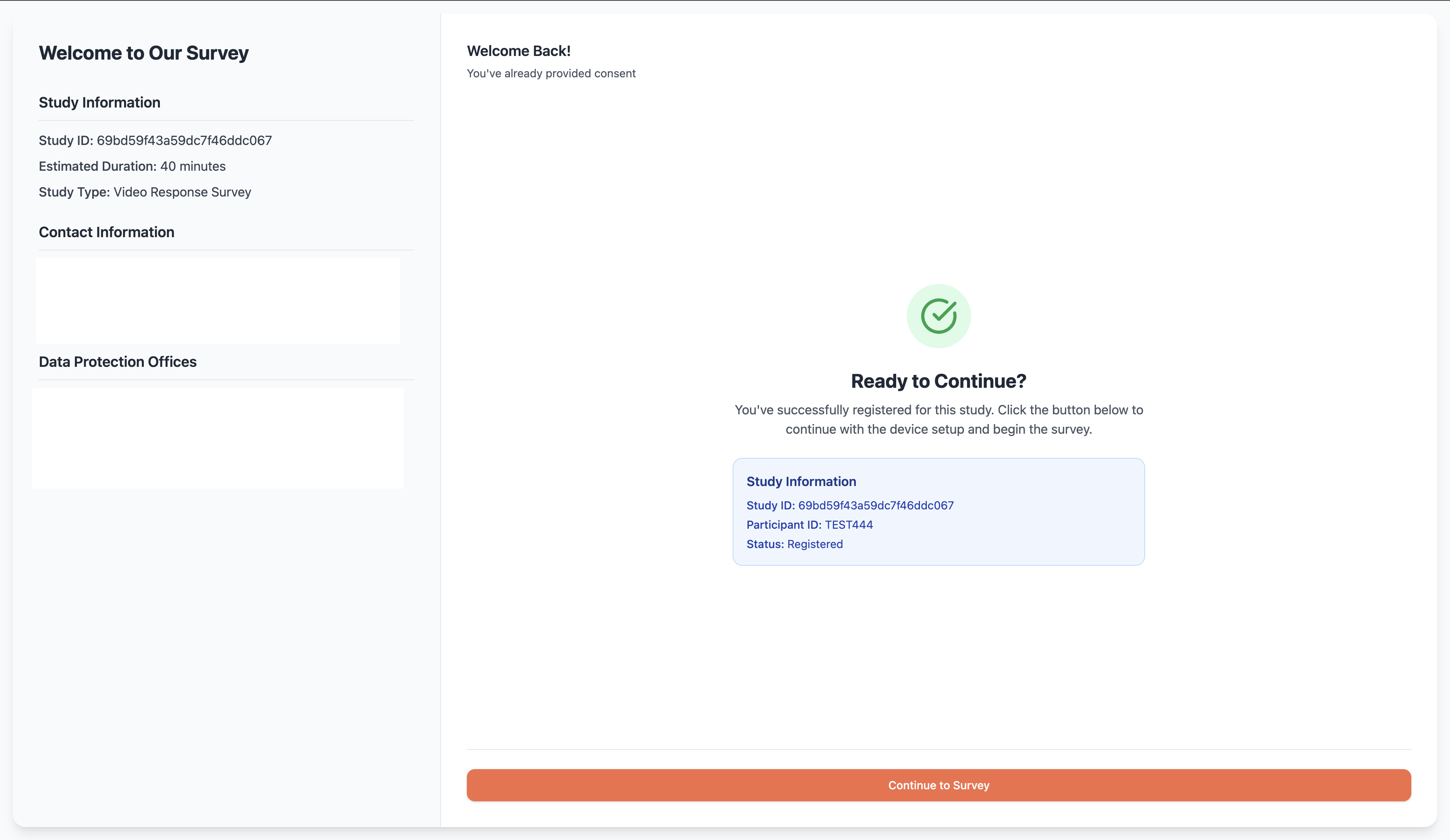}
        \caption{Registration completed.}
    \end{subfigure}

    \vspace{0.5em}

    \begin{subfigure}[t]{0.49\textwidth}
        \centering
        \includegraphics[width=\linewidth]{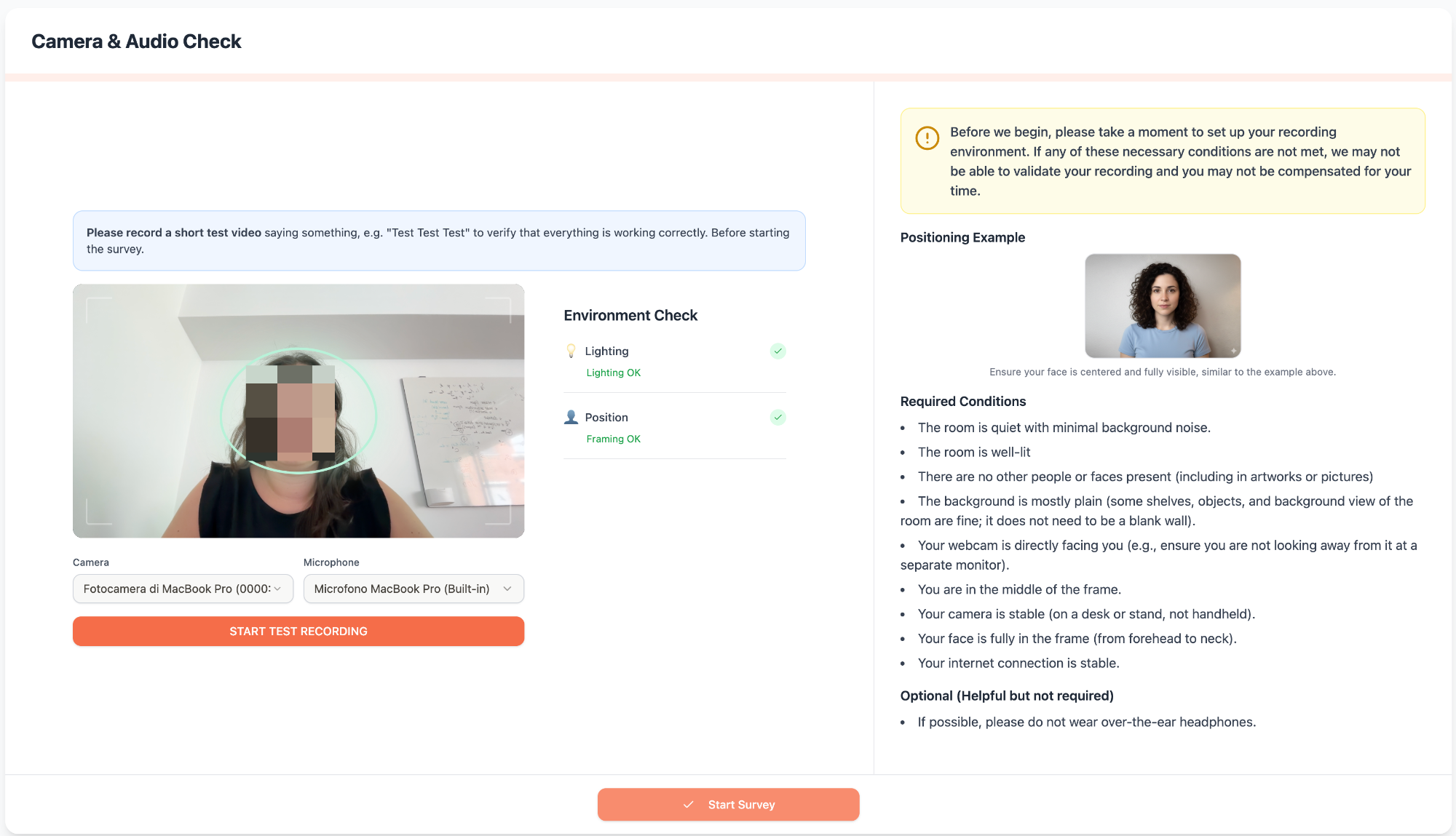}
        \caption{Camera and audio setup with environment verification.}
    \end{subfigure}
    \hfill
    \begin{subfigure}[t]{0.49\textwidth}
        \centering
        \includegraphics[width=\linewidth]{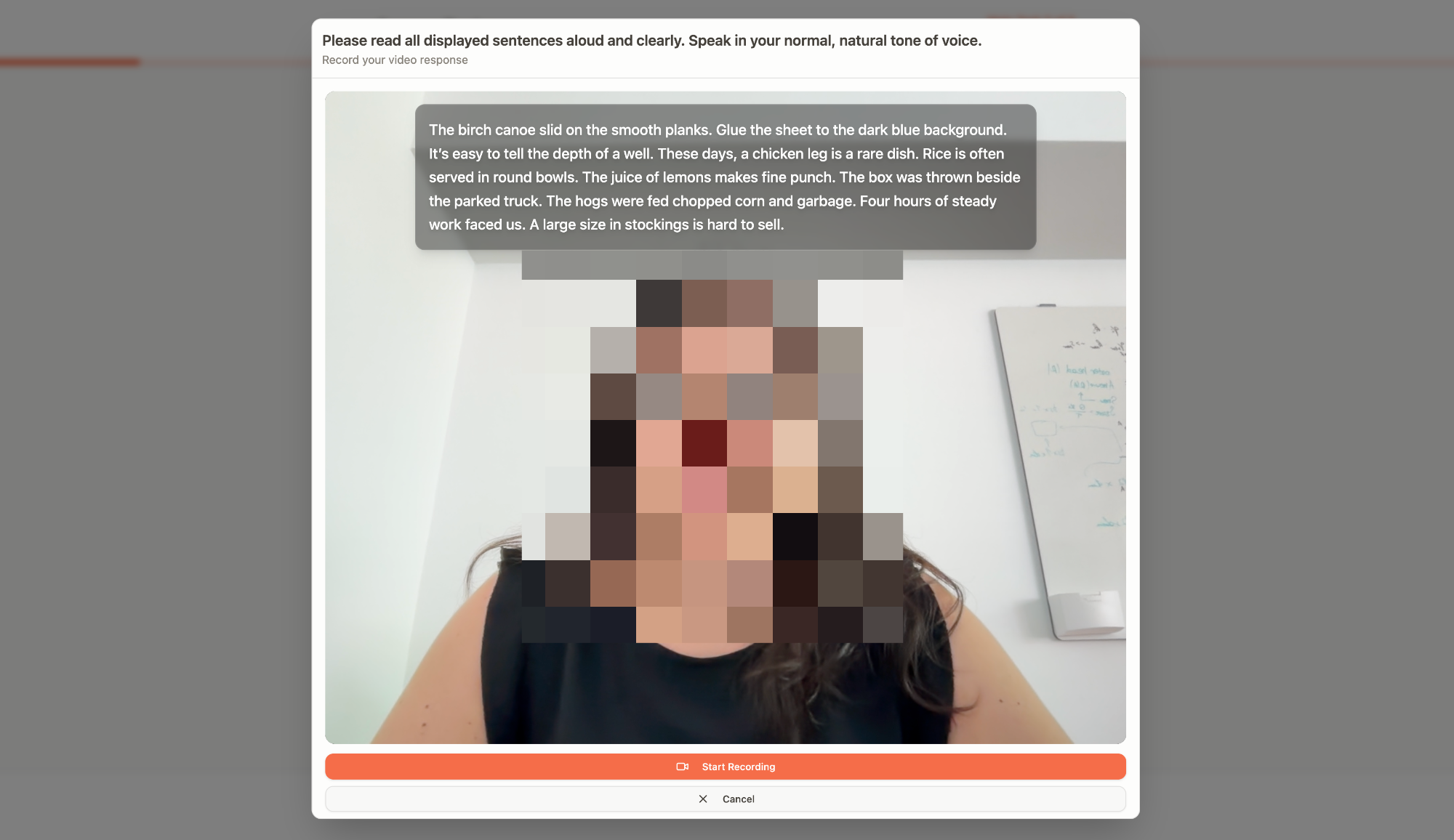}
        \caption{Browser-based recording interface.}
    \end{subfigure}

    \caption{Overview of the custom web application used to collect \ourname.
    Participants first review the study information and provide informed consent
    (a), after which, they can begin the survey (b). Before
    recording, the platform verifies the recording environment by checking the
    camera, microphone, framing, and lighting (c). Finally, participants complete the recording tasks through a browser-based interface that presents the prompts and
    captures synchronized audio and video (d).}
    \label{fig:webapp}
\end{figure*}

Figure~\ref{fig:webapp} illustrates the main stages of the application. Participants first review the study information and provide informed consent (Fig.\ref{fig:webapp} a). Once registered, they proceed to the recording workflow (Fig.\ref{fig:webapp} b), where the application verifies the recording environment by checking the camera, microphone, framing, and lighting while presenting the required recording conditions (Fig.\ref{fig:webapp} c). Finally, the platform presents the recording tasks - including standardized reading, spontaneous speech, and action-based prompts - through a unified browser interface that records audio and video for subsequent processing (Fig.\ref{fig:webapp} d).

\section{Real Data Acquisition}
\label{app:real_data}

\subsection{Participant Recruitment}

We collected the dataset in four stages: one pilot study and three subsequent collection phases. 
During the pilot study, we selected participants solely on the basis of their native or fluent language, without any additional demographic constraints. To increase accent diversity relative to existing \DF\ datasets, our recruitment focuses on speakers living outside the United States. 
Following the pilot study, we refined the screening criteria. First, we required the study’s target language to match the participant’s primary language and introduced demographic-balancing targets. The goal was to recruit approximately 48\% male, 48\% female, and 4\% non-binary participants. We largely achieved these targets across all language groups. However, recruiting non-binary participants proved more challenging for some languages. The French and German cohorts fell short of the target by one and two participants, respectively. 
For the second and third English collection phases, we restricted recruitment to native English speakers living in the United Kingdom, the United States, Ireland, Australia, Canada, and New Zealand. 

All participants were recruited through Prolific and compensated at the same rate of \pounds9/hour (recommended by the Prolific platform), irrespective of region, as payments were administered through the UK-based platform.

\subsection{Recording Environment}

Data collection was handled through a custom web-based application hosted on institutional infrastructure, and all recordings were stored on secure local servers.
Participants complete the study using their own devices in their usual environments. Before recording, participants were instructed to ensure that: 

\begin{itemize}
    \item The recording environment is quiet and free from excessive background noise; 
    \item The scene is adequately clean;
    \item No other people or visible faces appear in the frame;
    \item The webcam sits directly in front of the participant;
    \item The participant's face remains fully visible within the frame;
    \item The camera stays stable during recording;
    \item The internet connection is stable enough to support uninterrupted recording.
\end{itemize}

Unlike laboratory-controlled datasets, we collect recordings in unconstrained home environments. This setup introduces natural variability in the background, lighting, microphones, cameras, and overall recording quality. 

\subsection{Prompt Sources and Selection}
\label{app:prompt_source}
\paragraph{Standardized Speech Material} The voice cloning component of \ourname relies on two forms of standardized speech material. First, participants record a language-specific version of The North Wind and the Sun~\cite{aesop_northwind}. Phonetic and linguistic researchers use this passage widely, and it appears in the Handbook of the International Phonetic Association~\cite{InternationalPhoneticAssociation_1999}. Because it is available in multiple languages and offers broad phonetic coverage, it is well-suited to cross-language speaker analysis and voice cloning.

Alongside this paragraph, participants record a set of phonetically balanced sentences. Following the design philosophy of DeepSpeak~\cite{DeepSpeak}, which uses Harvard Sentences for English~\cite{english_elict}, \ourname uses language-specific, Harvard-style sentence collections or equivalent phonetically balanced corpora for each language~\cite{italian_elict,french_elict,spanish_elict,german_elict}. The sentence sets give us controlled speech samples with broad phonetic coverage while keeping a natural linguistic structure. 

\paragraph{Emotional and Randomized Reading Prompts} These prompts come from the M-AILABS corpus~\cite{MAILABS_2017}, which provides a large collection of multilingual text, allowing for the generation of diverse prompts across different languages and speaking styles. 
To improve consistency across recordings, we filter out candidate sentences with estimated reading times under 7 seconds or over 12 seconds. The remaining prompts take about 10 seconds to read on average. 
Additionally, to keep content safe, we run all the candidate sentences through a hate-speech detection model~\cite{hatespeech}. Following this step, we drop those sentences flagged as potentially harmful. 

\paragraph{Phonetic Diversity Selection} After filtering, we convert the remaining sentences into phoneme sequences and rank them using a phonetic richness score. We compute the score as $ 0.5 \cdot N_{\text{unique}} + 0.3 \cdot N_{\text{total}} + 0.2 \cdot L$,  where $N_{\text{unique}}$ denotes the number of unique phonemes, $N_{\text{total}}$ the total number of phonemes, and $L$ the sentence length. 
Sentences were subsequently clustered using TF–IDF~\cite{tfidf} representations of their phoneme sequences, allowing phonetically similar utterances to be grouped together and promoting diversity in the final selection.

\paragraph{Emotion Prompts} To build the emotion reading tasks, we employ sentiment classification on the chosen sentences, sorting each prompt into a positive, neutral, or negative sentiment category. 
We use the sentiment labels as a coarse filtering heuristic for the target emotion prompts (happy, sad, surprised, and neutral). We pull happy prompts mostly from the positive sentiment group, and sad prompts from the negative group. Surprised prompts come from the remaining positive samples, and neutral prompts come strictly from the neutral category. If we run short on prompts for a specific category, we draw extra sentences from the neutral pool. 

The final prompt set combines broad phonetic coverage, linguistic diversity, emotional variety, and content safety,providing a diverse basis for multilingual audio-visual data collection.

\section{Textual Prompts}
\label{app:textual_prompts}

Below, we report the textual prompts used during data collection together with the rationale behind the different task categories. For clarity, we show the English version of the prompts; the same task structure is used across all selected languages.

Table~\ref{tab:appendix_elicitation_sentences} reports the standardized elicitation sentences used for controlled reading. These prompts provide short, phonetically diverse utterances under relatively constrained speaking conditions. Table~\ref{tab:appendix_quick_fire_questions} contains short quick-fire questions designed to elicit brief spontaneous responses, while Table~\ref{tab:appendix_unscripted_questions} reports the open-ended questions used to obtain longer and more natural unscripted speech.

To complement these conversational tasks with longer controlled reading material, Table~\ref{tab:appendix_standard_sentences} lists excerpts from the public-domain novel \emph{Wives and Daughters} by Elizabeth Cleghorn Gaskell. These passages were used to collect extended read speech while avoiding copyright restrictions.

Table~\ref{tab:appendix_action_prompts} reports the action prompts, which combine speech with controlled head movements, gaze changes, facial expressions, hand gestures, and other visible actions. These prompts were introduced to increase the diversity of facial motion, pose, occlusion, and expression present in the recordings.

Finally, Table~\ref{tab:appendix_tasks} summarizes the overall task structure and the corresponding instructions shown to participants in the recording interface. It provides the mapping between the individual prompt types and the higher-level stages of the data-collection procedure.

\begin{table*}[t]
\centering
\small
\begin{tabularx}{\textwidth}{@{}rY rY@{}}
\toprule
\textbf{\#} & \textbf{Sentence} & \textbf{\#} & \textbf{Sentence} \\
\midrule
1 & The birch canoe slid on the smooth planks. & 2 & Glue the sheet to the dark blue background. \\
3 & It's easy to tell the depth of a well. & 4 & These days, a chicken leg is a rare dish. \\
5 & Rice is often served in round bowls. & 6 & The juice of lemons makes fine punch. \\
7 & The box was thrown beside the parked truck. & 8 & The hogs were fed chopped corn and garbage. \\
9 & Four hours of steady work faced us. & 10 & A large size in stockings is hard to sell. \\
\bottomrule
\end{tabularx}
\caption{Standardized elicitation sentences used for controlled reading.}
\label{tab:appendix_elicitation_sentences}
\end{table*}

\begin{table*}[t]
\centering
\small
\begin{tabularx}{\textwidth}{@{}rY rY@{}}
\toprule
\textbf{\#} & \textbf{Question} & \textbf{\#} & \textbf{Question} \\
\midrule
1 & What is your favorite color? & 2 & What is your favorite song or type of music? \\
3 & What did you do first thing this morning? & 4 & If you could live anywhere in the world, where would it be? \\
5 & What is something you are really good at? &  &  \\
\bottomrule
\end{tabularx}
\caption{Quick-fire questions used for short, spontaneous responses.}
\label{tab:appendix_quick_fire_questions}
\end{table*}

\begin{table*}[t]
\centering
\small
\begin{tabular}{ll}
\hline
\textbf{Type}        & \textbf{Question}                                              \\ \hline
Procedural           & Please describe the steps of your morning routine.             \\
Personal narrative   & Tell me about a time you helped someone, or someone helped you \\
Emotional reflection & What is something that always makes you feel happy?            \\
Opinion              & What do you think makes a good friend?                         \\ \hline
\end{tabular}
\caption{Open-ended unscripted questions used for longer spontaneous responses.}
\label{tab:appendix_unscripted_questions}
\end{table*}

\begin{table*}[t]
\centering
\begin{tabularx}{\textwidth}{@{}X@{}}
\toprule
\textbf{Excerpt} \\        
\midrule
but it was such an unheard-of thing for him to be sitting in that room in the middle of the day, reading or making pretence to read, that she had never thought of his remaining.        \\
Lady Cumnor and her daughters received all the school visitors at the Towers, the great family mansion standing in aristocratic seclusion in the center of the large park,               \\
But every time that Missis Gibson was struck by Cynthia's beauty, she thought it more and more advisable that Mister Osborne Hamley should be cheered up by a quiet little dinner-party. \\
But they had understood that he was not coming to the Hall until the following week, and therefore they had felt themselves at full liberty this afternoon to follow their own devices.  \\
Then the carriage came round, and after numberless last words from the earl - who appeared to have put off every possible direction to the moment when he stood,                         \\
Molly made a point of turning the conversation from all personal subjects after this, and kept the Squire talking about the progress of his drainage during the rest of lunch.           \\
His daughter-in-law came in at the sound, and told the Squire that he had these coughing-bouts very frequently, and that she thought he would go off in one of them before long.         \\
She has brought up the young ladies at the Towers, and has a daughter of her own, therefore it is probable she will have a kind, motherly feeling towards Molly.                         \\
The small window that gave it light looked right on to the moor, as it was called; and by day the check curtain was drawn aside so that he might watch the progress of the labour.       \\
and although this attention on her part was melting the heart he tried to harden, he controlled himself into writing her the briefest acknowledgments.                                   \\ \hline
\bottomrule
\end{tabularx}
\caption{Public-domain passages from Wives and Daughters~\cite{gaskell1866wives} by Elizabeth Cleghorn Gaskell used for extended controlled reading.}
\label{tab:appendix_standard_sentences}
\end{table*}

\begin{table*}[t]
\centering
\scriptsize
\centering

\small

\begin{tabularx}{\textwidth}{@{}p{0.25\textwidth}X@{}}

\toprule

\textbf{ID} & \textbf{Instruction} \\

\midrule

Lean forward and emphasize point &

[LEAN SLIGHTLY FORWARD TOWARDS THE CAMERA] while saying: 

``Let me emphasize this, this is really important:'' 

[Wait approx. 1 second] then emphasize: 

``Pineapple on pizza is good!'' 

[RETURN TO THE STARTING POSITION]. \\

Look down to the right &

[LOOK DOWN TO THE RIGHT] as if you were checking something there, and say: 

``If I look here, I can see what's happening on my right side.'' \\

Look down to the middle and focus &

[LOOK SLIGHTLY DOWN TO THE MIDDLE] Pretend you are checking a document. 

Name any date and briefly say that the document looks correct. \\

Look down to the left &

[LOOK DOWN TO THE LEFT], as if you were checking something there, and say: 

``If I look here, I can see what's happening on my left side.'' \\

Look up to the left and recall &

[LOOK UP TO THE LEFT], as if you were remembering something, and say: 

``Looking up here reminds me of the pattern on the ceiling at home.'' \\

Look up to the middle and examine &

[LIFT YOUR GAZE UPWARDS], as if you were examining something above you, and say: 

``I'm just looking up to see how everything fits together.'' \\

Look up to the right and notice &

[LOOK UP TO THE RIGHT], as if you were noticing something new, and say: 

``Oh, I can really see that corner now---I hadn't noticed that before.'' \\

Wave hand in front of face &

[WAVE YOUR HAND LOOSELY FROM SIDE TO SIDE IN FRONT OF YOUR FACE], while 

describing the motion: 

``You see how my hand moves, back and forth, back and forth.'' \\

Pretend to yawn &

[PRETEND TO YAWN] while saying: ``I'm a little tired today.'' 

[OPTIONAL ACTION: COVER YOUR MOUTH WITH YOUR HAND] \\

Pretend to laugh &

[LAUGH BRIEFLY NATURALLY OR PLAYFULLY] while saying: 

``That was hilarious!'' \\

Clap continuously, slowly, and loudly (5 sec.) &

[CLAP CONTINUOUSLY SLOWLY, LOUDLY AND AUDIBLY FOR ABOUT FIVE (5) SECONDS]. 

Then say: ``That's a start. But there is still a lot to do.'' \\

Shake head twice to say no &

[SHAKE YOUR HEAD TWICE CLEARLY FROM SIDE TO SIDE] while rejecting an idea: 

``No, I really don't think that's the right approach.'' \\

Nod twice to say yes &

[NOD NATURALLY TWICE] while agreeing: 

``Yes, absolutely, that makes perfect sense to me.'' \\

Tilt head to the right and think &

[TILT YOUR HEAD STRONGLY TO YOUR RIGHT SHOULDER WHILE THINKING] 

``Hmm... let me think about that for a moment\ldots'' 

[RETURN TO THE STARTING POSITION] and then say: 

``Okay, here's what I think.'' \\

Tilt head to the left and think &

[TILT YOUR HEAD STRONGLY TO YOUR LEFT SHOULDER WHILE THINKING] 

``I might see this differently... let me consider\ldots'' 

[RETURN TO THE STARTING POSITION] and then say: 

``Yes, that's one way to see it.'' \\

Smile widely &

[SMILE OPENLY AND HEARTILY] while talking about something nice: 

``The surprise party was honestly the nicest moment in years!'' \\

Scan the room from left to right &

[SCAN THE ROOM SLOWLY FROM LEFT TO RIGHT] Briefly name and describe three 

things (real or imaginary) that you see. Take about 10 seconds to do this. \\

Describe object in hand &

[TAKE ANY OBJECT AND HOLD IT VISIBLY TOWARDS THE CAMERA] Look at it closely 

and describe its details (e.g., shape, color, or material) for 15 seconds. \\

\bottomrule

\end{tabularx}
\caption{Action prompts used during data collection.}
\label{tab:appendix_action_prompts}
\end{table*}

\begin{table*}[t]
\centering
\small\begin{tabularx}{\textwidth}{llX}

\hline

\textbf{Type} & \textbf{Title} & \textbf{Description / Messages} \\ \hline

Unscripted Answers

& 

& Please answer the question in detail (target: approx. 30 seconds). \\

Voice Cloning Prompts

& Reading a short list of sentences

& Please read all displayed sentences aloud and clearly. Speak in your normal, natural tone of voice. \\

Voice Cloning Prompts

& Reading a sample text

& Read the entire text aloud. Make sure to speak calmly and clearly -- just as you would in everyday life. \\

Q\&A Task

& Question and answer interview

& You will see a list of questions. Read each question aloud, then answer it immediately in the same recording. \\

Interleave Reading Task &

Mixed Reading and Action Tasks &

You have successfully completed the introductory tasks. Now we will move on to the main block of data entry. In this section, different types of tasks will alternate:

\begin{itemize}[leftmargin=*,nosep]

    \item Sometimes you will read sentences in a neutral tone.

    \item Sometimes you will read sentences with a given emotion.

    \item Some tasks combine speaking with small movements or facial expressions.

\end{itemize} \\

Standardized Reading Task

& Reading standardized sentences

& Please read the displayed sentence aloud. Try to speak naturally. \\

Randomized Reading Task

& Speaking with Emotion

& Read the following sentence aloud. Try to express the indicated emotion. \\

Action Prompts

& Action prompts with speech

& Please read the instruction and then perform the described action while speaking the provided text. \\

Introduction Questions

& Open-ended Questions

& To begin, we will show you four open-ended questions. There are no right or wrong answers. You may also make something up if you like. \\ \hline

\end{tabularx}

\caption{Task structure and interface messages used during data collection.}
\label{tab:appendix_tasks}
\end{table*}

\section{Generation Methods}
\label{app:generation_techniques}
\subsection{Audio Models}

For audio generation, we employ four multilingual text-to-speech models: Qwen3-TTS~\cite{Qwen}, FishAudio2~\cite{fishaudio}, OmniVoice~\cite{omnivoice}, and ChatterBox~\cite{chatterbox}.
\textbf{Qwen3-TTS} is an open-source multilingual text-to-speech model that supports zero-shot voice cloning from short reference audio samples. The model captures speaker-specific characteristics, such as timbre and speaking style, and generates novel utterances that preserve the target speaker's vocal identity. \textbf{FishAudio2} provides rapid voice cloning from short reference recordings, typically requiring only 10-30 seconds of speech. It can synthesize arbitrary utterances while maintaining speaker characteristics and natural prosody, enabling realistic speaker-consistent speech generation. \textbf{OmniVoice} is a multilingual text-to-speech model that combines language-agnostic speech representations with diffusion language models to achieve high-fidelity speech synthesis and speaker transfer. Given a short reference recording, it generates arbitrary utterances while preserving speaker-specific attributes, including vocal timbre and prosodic patterns. \textbf{ChatterBox} supports zero-shot voice cloning and expressive speech synthesis from short reference audio. Beyond preserving speaker identity and speaking style, the model provides controllable emotional expressiveness, enabling diverse speech realizations from the same speaker.

We leverage the zero-shot voice-cloning capabilities and the multilingual nature of these models to generate realistic audio DeepFakes. The synthesized speech is subsequently combined with video generation engines to produce the final audio-visual fake samples. 

\subsection{Video Models}
Following the recent literature, we consider three categories of video manipulation: \emph{face swapping}, \emph{lip synchronization}, and \emph{portrait animation}. Face swapping replaces the identity of a target subject while preserving facial motion; lip synchronization alters mouth movements to match a target speech signal without changing the speaker's identity; and avatar/talking-head animation synthesizes an entire facial performance from a reference portrait or avatar.

\paragraph{FaceSwap} Face swapping replaces the identity of the target subject while preserving facial dynamics, including head pose, facial expressions, and eye movements. \textbf{CanonSwap}~\cite{canonswap} performs identity transfer in a canonical facial space, explicitly disentangling identity from facial motion. After removing motion-related information, the source identity is injected into the canonical representation, thereby restoring the original facial dynamics and improving temporal consistency while preserving expressions, head pose, and lip synchronization. 
\textbf{DeepLiveCam}~\cite{deeplivecam} is a real-time face swapping framework designed for both prerecorded videos and live webcam streams. It combines robust face detection, identity replacement, and face enhancement to generate temporally coherent face-swapped videos with minimal user intervention.
\textbf{HiFiFace}~\cite{HifiFace} leverages a 3D shape-aware identity representation to preserve the geometric characteristics of the source face during identity transfer. Combined with a Semantic Facial Fusion module, it adaptively merges source identity and target facial attributes, producing realistic face swaps while maintaining pose, expression, and illumination consistency.
\textbf{InSwapper}~\cite{inswapper} uses InsightFace to detect and localize facial landmarks, aligns the source and target faces, and then superimposes the transformed source identity onto each target frame.

Both HiFiFace and InSwapper are generated through the FaceFusion Library~\cite{facefusion}. 
\paragraph{LipSynchronization} Lip synchronization modifies only the mouth region to match a target speech signal while preserving the speaker's identity, facial appearance, and the remaining facial dynamics. \textbf{KeySync}~\cite{keysync} is a two-stage lip synchronization framework designed to produce high-resolution, temporally consistent talking-face videos. The method explicitly addresses two major challenges of lip synchronization: expression leakage from the source video and facial occlusions. It introduces a masking strategy that suppresses undesired facial expressions while preserving identity and background information, resulting in realistic lip movements with improved temporal consistency and robustness to occlusions. \textbf{LatentSync}~\cite{latentsync} is an end-to-end audio-conditioned latent diffusion framework. It performs lip-sync generation directly in the latent space without relying on intermediate motion representations. To improve temporal consistency, it introduces a Temporal REPresentation Alignment (TREPA) module that aligns temporal video features across consecutive frames, while enhanced SyncNet supervision strengthens audio-visual synchronization. The resulting videos exhibit accurate lip movements, high visual quality, and temporal coherence. \textbf{MuseTalk}~\cite{musetalk} is a real-time audio-driven lip synchronization framework that generates high-quality talking-face videos by performing latent space inpainting on the mouth region. The method leverages a pretrained latent diffusion model, along with audio and visual encoders, to synthesize realistic lip movements while preserving the target identity, facial expressions, and head pose. Its latent-space formulation enables efficient inference while maintaining high visual fidelity and accurate audio-visual synchronization.
\paragraph{Portrait Animation} Portrait animation synthesizes an entire facial performance from a reference portrait and a driving signal, such as audio or video. Unlike lip synchronization, these methods jointly generate lip movements, facial expressions, head pose, and, in some cases, upper-body motion, while preserving the identity of the reference subject. \textbf{AvatarForcing}~\cite{avatarforcing} is a diffusion-based talking-head generation model that animates a reference portrait from an input speech signal. By employing a one-step streaming diffusion framework, it generates photorealistic talking-head videos with synchronized lip movements, stable facial identity, and temporally consistent facial dynamics, enabling high-quality long-form avatar generation.
\textbf{HunyuanPortrait}~\cite{hunyuanportrait} is a diffusion-based portrait animation model that animates a reference image using motion extracted from a driving video. It explicitly disentangles identity and motion through separate appearance and motion representations, enabling accurate lip synchronization while preserving facial identity and generating realistic facial expressions, head movements, and upper-body dynamics.
\textbf{PersonaLive}~\cite{personalive} is a diffusion-based portrait animation model that synthesizes expressive talking-head videos from a reference image. It leverages hybrid motion representations to jointly model facial expressions and head movements while preserving the subject’s identity, and adopts a streaming diffusion framework to generate temporally consistent animations with low latency.  
\textbf{UniAVGen}~\cite{uniavgen} is a unified audio-visual generation model that jointly synthesizes synchronized speech and talking-head videos using dual diffusion transformers. Through asymmetric cross-modal interactions and face-aware modulation, the model explicitly couples audio and visual generation to improve lip synchronization, temporal consistency, and identity preservation. In our pipeline, we employ UniAVGen for talking-head animation from speech.

\paragraph{Implementation Details} All generation methods are employed out of the box, using their publicly available implementations. For each method, we follow the released code and the corresponding preprocessing and inference pipelines, including the required input preparation, face detection and alignment, audio preprocessing, and post-processing steps, whenever applicable. We do not introduce modifications to the underlying architectures or generation procedures. This setup is intended to reflect each method’s publicly released generation pipeline, support reproducibility, and provide a representative evaluation of contemporary DeepFake generation techniques. 

\section{Raw Data Processing and Audio-Visual Curation}
\label{app:preprocessing_fakes}

\subsection{Real Data Preprocessing and Standardization}
\label{app:preprocessing}

\paragraph{Video and Audio Encoding.}
Raw recordings captured via the web-based acquisition interface are saved in WebM format (VP8/VP9 video, Opus audio). We batch-transcode all raw streams to MP4 using FFmpeg, re-encoding video tracks with H.264 (preserving native resolution) and audio tracks with AAC at 48\,kHz. Constant frame rates (CFR) are estimated per clip based on decoded frame counts and total duration; whenever timestamps are non-uniform, videos are processed in variable-frame-rate (VFR) mode to eliminate motion jitter and temporal distortion.

\paragraph{Transcript Extraction and Refinement.}
Audio transcripts are extracted using Whisper Large~\cite{whisper}. Because these transcripts exclusively serve as intermediate text prompts for TTS voice-cloning engines (rather than evaluation targets), word-for-word ground-truth precision is secondary to phonetic naturalness. We utilize Gemini 2.5~\cite{gemini} in a zero-shot prompt setup to correct common ASR acoustic artifacts, restore punctuation, and clean up broken sentence boundaries, providing coherent text inputs for downstream TTS synthesis.

\paragraph{Audio-Visual Temporal Alignment.}
Audio-driven lip-synchronization and talking-head models require precise alignment between the source video and the conditioning audio. Audio streams fed to synthesis pipelines are normalized and resampled to 16\,kHz. When the source video duration exceeds the driving audio by $\Delta t > 0.5$\,s, we trim the terminal video frames to match the audio. If the driving audio exceeds the source video by $\Delta t > 0.5$\,s, the pair is discarded rather than looped or frozen, preventing synthetic temporal discontinuities from leaking into downstream detector benchmarks.

\subsection{Best-Eye-Pose Computation}
\label{app:best_eye_pose}
Facial landmarks are detected in each frame using the MediaPipe Face Mesh~\cite{kartynnik2019realtimefacialsurfacegeometry}. Eyes' openness was quantified using the Eye Aspect Ratio (EAR)~\cite{Soukupov2016RealTimeEB}, computed for each eye and then averaged. We estimated head orientation from six 2D facial landmarks and their 3D facial coordinates by using the PnP (perspective-n-point) implementation from the OpenCV library~\cite{opencv_library}. The retained frames have an average EAR of at least 0.22 and the estimated pose $|\mathrm{yaw}| \leq 12^\circ$ - so the estimated left-right rotation of the participant's head around the vertical axis-, pitch deviation $\leq 12^\circ$, and $|\mathrm{roll}| \leq 10^\circ$. Among the retained frames, we selected the one minimizing a normalized pose-deviation score.

\section{Benchmark Evaluation Details}

\subsection{Dataset Preprocessing and Model Training Setup}
\label{app:exp_setup}
\textbf{Dataset Preprocessing} For each evaluated model, we preprocess the audio and video inputs according to the requirements of the corresponding model and its official implementation. This ensures that each method is evaluated using the input representation and preprocessing procedure for which it was originally designed. 

\textbf{Model Setup} While SpeechForensics~\cite{speechforensics} is evaluated with the pretrained model provided by the authors, we train task-specific models for AVH-Align~\cite{AVH-Align} and \bn~\cite{braven}.
For AVH-Align, we train the supervised version using the code released by the authors, without modifying any hyperparameters.

For \bn, instead, we use the pretrained Base+ audio and video encoders as frozen feature extractors and train a binary linear probe on top of their representations. The encoders remain frozen throughout training. For each 4-second input clip, we temporally average the encoder outputs to obtain a 768-dimensional representation for each modality. The unimodal classifiers consist of a single 768-to-1 linear layer, while the audiovisual classifier operates on the concatenated 1536-dimensional audio-video representation. We train the linear probes for 10 epochs using AdamW~\cite{adamw} (learning rate ($10^{-3}$), weight decay ($10^{-4}$)) and binary cross-entropy~\cite{bce} with logits. 

\subsection{Training and testing on \ourname}

We additionally evaluate the models when trained on \ourname training data and tested on the test set. Results are reported in Tab.~\ref{tab:babel_training}. AVH-Align achieves the best performance, with an AUC of 97.37, outperforming \bn-AV 95.52, \bn-V (93.47), and \bn-A (80.03). Overall, we confirm that the test set is ``solvable'' when training and testing in-domain.

\begin{table}[t]
    \centering
    \begin{tabular}{lc}
        \toprule
        \textbf{Model} & \textbf{AUC} \\
        \midrule
        \bn-A  & 80.03 \\
        \bn-V  & 93.47 \\
        \bn-AV & 95.52 \\ \hline
        AVH-Align & \textbf{97.37} \\
        \bottomrule

    \end{tabular}
     \caption{AUC results when training and evaluating on \ourname (test set). Best performance is highlighted in bold.}
    \label{tab:babel_training}
\end{table}

\subsection{~\ourname Results Breakdown}

\begin{table*}[ht]

\centering

\small

\resizebox{\linewidth}{!}{%

\begin{tabular}{ll|c|ccccc|ccccc|c|c}

\hline

\textbf{Model} &

\textbf{\begin{tabular}[c]{@{}l@{}}Training\\ Data\end{tabular}} &

\multicolumn{1}{c|}{\textbf{AUC}} &

\multicolumn{5}{c|}{\textbf{AvatarForcing}} &

\multicolumn{5}{c|}{\textbf{PersonaLive}} &

\multicolumn{1}{c|}{\textbf{\begin{tabular}[c]{@{}c@{}}Hunyuan\\ Portrait\end{tabular}}} &

\multicolumn{1}{c}{\textbf{\begin{tabular}[c]{@{}c@{}}UniAV\\ Gen\end{tabular}}}

\\ \hline

& & \multicolumn{1}{c|}{} &

\rotatebox{90}{ChatterBox} &

\rotatebox{90}{FishAudio2} &

\rotatebox{90}{OmniVoice} &

\rotatebox{90}{Qwen3-TTS} &

\multicolumn{1}{c|}{\rotatebox{90}{Real Audio}} &

\rotatebox{90}{ChatterBox} &

\rotatebox{90}{FishAudio2} &

\rotatebox{90}{OmniVoice} &

\rotatebox{90}{Qwen3-TTS} &

\multicolumn{1}{c|}{\rotatebox{90}{Real Audio}} &

\multicolumn{1}{c|}{\rotatebox{90}{Real Audio}} &

\rotatebox{90}{Real Audio}

\\ \hline

\textbf{\bn-A} & DeepSpeak & 60.09 &

65.25 & 68.44 & 67.27 & 67.91 & 49.79 &

65.92 & 68.99 & 70.13 & 68.90 & 51.70 &

53.58 & 38.05 \\

\textbf{\bn-V} & DeepSpeak & 70.31 &

75.62 & 79.77 & 74.50 & 77.44 & 63.73 &

66.95 & 67.58 & 67.01 & 67.10 & 67.96 &

66.51 & 54.10 \\

\textbf{\bn-AV} & DeepSpeak & {\ul 72.05} &

74.10 & 77.77 & 74.17 & 75.59 & 61.51 &

81.91 & 81.62 & 81.83 & 81.22 & 72.11 &

66.84 & 52.13 \\ \hline

\textbf{\bn-A} & MAVOS-DD & 43.93 &

34.22 & 38.58 & 45.78 & 40.17 & 51.89 &

33.01 & 37.39 & 43.95 & 38.21 & 48.32 &

48.97 & 65.72 \\

\textbf{\bn-V} & MAVOS-DD & 72.56 &

72.61 & 75.84 & 76.57 & 75.40 & 73.03 &

71.82 & 72.45 & 71.70 & 71.74 & 72.06 &

66.88 & 69.90 \\

\textbf{\bn-AV} & MAVOS-DD & {\ul 73.88} &

74.12 & 78.14 & 78.75 & 77.27 & 74.62 &

72.32 & 73.88 & 73.62 & 72.83 & 72.65 &

68.90 & 68.51 \\ \hline

\textbf{AVH-Align} & AV1M & 18.51 &

3.86 & 4.84 & 6.17 & 4.46 & 5.04 &

35.36 & 37.42 & 29.94 & 32.11 & 19.07 &

35.56 & 28.84 \\ \hline

\textbf{AVH-Align sup} & MAVOS-DD & 67.11 &

73.65 & 65.83 & 66.38 & 70.90 & 55.36 &

82.50 & 74.98 & 70.06 & 74.39 & 65.42 &

73.15 & 62.23 \\

\textbf{AVH-Align sup} & FakeAVCeleb & {\ul 71.37} &

78.87 & 75.20 & 80.01 & 75.89 & \textbf{84.24} &

72.31 & 67.85 & 73.01 & 70.03 & \textbf{82.68} &

\textbf{70.81} & 47.27 \\

\textbf{AVH-Align sup} & AV1M & 41.26 &

57.11 & 12.25 & 38.29 & 15.97 & 27.92 &

74.41 & 15.10 & 33.42 & 18.38 & 51.97 &

50.16 & 40.41 \\ \hline

\textbf{SpeechForensics} & \begin{tabular}[c]{@{}l@{}}VoxCeleb2\\ \& LRS3\end{tabular}         & {\ul \textbf{79.79}} &

\textbf{85.67} & \textbf{85.67} & \textbf{85.06} & \textbf{85.38} & \textbf{75.71} &

\textbf{97.54} & \textbf{96.81} & \textbf{96.44} & \textbf{97.24} & \textbf{87.71} &

69.68 & \textbf{78.22} \\ \hline

\textit{Avg. Audio} & -- & 52.01 &

49.74 & 53.51 & 56.53 & 54.04 & 50.84 &

49.47 & 53.19 & 57.04 & 53.56 & 50.01 &

51.28 & 51.89 \\

\textit{Avg. Video} & -- & 71.44 &

74.12 & 77.81 & 75.54 & 76.42 & 68.38 &

69.39 & 70.02 & 69.36 & 69.42 & 70.01 &

66.70 & \textbf{62.00} \\

\textit{Avg. AV} & -- & \textbf{74.27} &

78.19 & 79.20 & 79.50 & 78.53 & 74.02 &

81.02 & 80.04 & 81.23 & 80.33 & 78.79 &

69.06 & 61.53 \\ \hline

\end{tabular}%

}

\caption{Fine-grained benchmark evaluation across talking-head and audio-visual generation methods decomposed by driving audio condition (Part 1). Results are reported in AUC (\%).}

\label{tab:app_eval_audiobreakdown_p1}

\end{table*}

\begin{table*}[ht]

\centering

\resizebox{\linewidth}{!}{%

\begin{tabular}{ll|c|ccccc|ccccc|ccccc|c|c|c|c}

\hline

\textbf{Model} &

\textbf{\begin{tabular}[c]{@{}l@{}}Training\\ Data\end{tabular}} &

\multicolumn{1}{c|}{\textbf{AUC}} &

\multicolumn{5}{c|}{\textbf{KeySync}} &

\multicolumn{5}{c|}{\textbf{MuseTalk}} &

\multicolumn{5}{c|}{\textbf{LatentSync}} &

\textbf{\begin{tabular}[c]{@{}c@{}}Canon\\ Swap\end{tabular}} &

\textbf{HiFiFace} &

\textbf{\begin{tabular}[c]{@{}c@{}}In\\ Swapper\end{tabular}} &

\textbf{\begin{tabular}[c]{@{}c@{}}DeepLive\\ Cam\end{tabular}} 

\\ \hline

& & \multicolumn{1}{c|}{} &

\rotatebox{90}{ChatterBox} &

\rotatebox{90}{FishAudio2} &

\rotatebox{90}{OmniVoice} &

\rotatebox{90}{Qwen3-TTS} &

\rotatebox{90}{Real Audio} &

\rotatebox{90}{ChatterBox} &

\rotatebox{90}{FishAudio2} &

\rotatebox{90}{OmniVoice} &

\rotatebox{90}{Qwen3-TTS} &

\rotatebox{90}{Real Audio} &

\rotatebox{90}{ChatterBox} &

\rotatebox{90}{FishAudio2} &

\rotatebox{90}{OmniVoice} &

\rotatebox{90}{Qwen3-TTS} &

\rotatebox{90}{Real Audio} &

\rotatebox{90}{Real Audio} &

\rotatebox{90}{Real Audio} &

\rotatebox{90}{Real Audio} &

\rotatebox{90}{Real Audio}

\\ \hline

\textbf{\bn-A} & DeepSpeak & 60.09 &

64.61 & 66.43 & 72.09 & \textbf{71.88} & 50.51 &

57.15 & 60.5 & 77.33 & 63.09 & 50.62 &

62.82 & 76.45 & 71.08 & 67.46 & 50.7 &

49.79 & 49.79 & 49.75 & 49.46 \\

\textbf{\bn-V} & DeepSpeak & 70.31 &

63.24 & 66.93 & 63.74 & 65.96 & 54.57 &

85.41 & \textbf{88.5} & \textbf{90.45} & \textbf{86.8} & \textbf{81.44} &

74.71 & 81.46 & 72.64 & 72.5 & 59.92 &

72.07 & 60.45 & 67.11 & 67.92 \\

\textbf{\bn-AV} & DeepSpeak & {\ul 72.05} &

62.98 & 67.24 & 67.8 & 67.05 & 53.42 &

79.80 & 81.75 & 89.81 & 81.35 & 73.87 &

74.51 & \textbf{81.84} & 76.15 & 72.54 & 57.57 &

69.49 & 58.92 & 65.81 & 65.96 \\ \hline

\textbf{\bn-A} & MAVOS-DD & 43.93 &

40.38 & 38.37 & 48.53 & 38.87 & 46.1 &

40.92 & 42.19 & 51.51 & 42.22 & 47.57 &

40.4 & 36.84 & 50.08 & 43.65 & 47.96 &

51.38 & 48.78 & 48.72 & 50.65 \\

\textbf{\bn-V} & MAVOS-DD & 72.56 &

58.43 & 59.32 & 62.44 & 59.37 & 56.44 &

76.62 & 76.88 & 80.73 & 77.38 & 77.41 &

71.47 & 73.29 & 75.11 & 76.71 & \textbf{60.75} &

79.92 & 69.82 & 74.67 & 76.6 \\

\textbf{\bn-AV} & MAVOS-DD & {\ul 73.88} &

60.39 & 61.48 & 66.13 & 61.88 & 56.85 &

78.72 & 79.7 & 84.78 & 80.06 & 78.91 &

72.95 & 74.53 & 77.21 & 77.59 & 60.49 &

81.51 & \textbf{69.57} & 75.05 & 77.3 \\ \hline

\textbf{AVH-Align} & AV1M & 18.51 &

5.68 & 12.08 & 10.61 & 7.82 & 4.56 &

31.17 & 29.29 & 26.55 & 29.93 & 6.52 &

14.05 & 8.18 & 15.63 & 12.65 & 6.56 &

36.03 & 9.88 & 9.29 & 39.42 \\ \hline

\textbf{AVH-Align sup} & MAVOS-DD & 67.11 &

65.73 & 43.71 & 47.61 & 46.49 & 36.08 &

80.73 & 71.06 & 76.67 & 74.45 & 59.95 &

82.23 & 77.19 & 68.59 & 71.99 & 43.82 &

78.6 & 57.31 & 66.16 & 71.8 \\

\textbf{AVH-Align sup} & FakeAVCeleb & {\ul 71.37} &

48.43 & 55.35 & 59.08 & 50.26 & \textbf{57.75} &

65.99 & 64.8 & 67.17 & 65.24 & 78.11 &

61.25 & 53.78 & 61.21 & 57.47 & 57.65 &

75.13 & 70.11 & 77.85 & 73.77 \\

\textbf{AVH-Align sup} & AV1M & 41.26 &

\textbf{73.83} & 16.80 & 33.02 & 18.36 & 25.43 &

77.34 & 18.45 & 35.5 & 20.77 & 29.86 &

75.7 & 23.04 & 39.97 & 20.3 & 34.43 &

49.96 & 77.95 & \textbf{78.34} & \textbf{50.35} \\ \hline

\textbf{SpeechForensics} &\begin{tabular}[c]{@{}l@{}}VoxCeleb2\\ \& LRS3\end{tabular}         & {\ul \textbf{79.79}} &

58.00 & \textbf{73.94} & \textbf{72.53} & 63.91 & 43.75 &

\textbf{85.53} & 83.12 & 86.1 & 84.7 & 69.86 &

\textbf{82.3} & 72.57 & \textbf{82.7} & \textbf{79.91} & 49.29 &

\textbf{80.22} & 63.62 & 70.98 & 72.71 \\ \hline

\textit{Avg. Audio} & -- & 52.01 &

52.50 & 52.40 & 60.31 & 55.38 & 48.31 &

49.04 & 51.35 & 64.42 & 52.66 & 49.10 &

51.61 & 56.65 & 60.58 & 55.56 & 49.33 &

50.59 & 49.29 & 49.24 & 50.06 \\

\textit{Avg. Video} & -- & 71.44 &

60.84 & 63.13 & 63.09 & 62.67 & 55.51 &

81.02 & 82.69 & 85.59 & 82.09 & 79.43 &

73.09 & 77.38 & 73.88 & 74.61 & 60.34 &

76.00 & 72.26 & 65.14 & 70.89 \\

\textit{Avg. AV} & -- & \textbf{74.27} &

57.45 & 64.50 & 66.39 & 60.78 & 52.94 &

77.51 & 77.34 & 81.97 & 77.84 & 75.19 &

72.75 & 70.68 & 74.32 & 71.88 & 58.57 &

75.38 & 66.20 & 72.90 & 72.34 \\ \hline

\end{tabular}%
}
\caption{Fine-grained benchmark evaluation across lip-synchronization and face-swapping manipulation methods decomposed by driving audio condition (Part 2). Results are reported in AUC (\%).}
\label{tab:app_eval_audiobreakdown_p2}
\end{table*}

\begin{table*}[ht]
\centering
\begin{tabular}{rlcccc}
\hline
\multicolumn{2}{l}{\textbf{Engine}} & \textbf{\begin{tabular}[c]{@{}c@{}} \# Rated\\ Samples\end{tabular}}          & \textbf{Mean Score} & \textbf{\begin{tabular}[c]{@{}c@{}}Deception Rate\\ (\%)\end{tabular}} & {\color[HTML]{F56B00} \textbf{\begin{tabular}[c]{@{}c@{}}AUC\\ (\%)\end{tabular}}} \\ \hline
\multicolumn{2}{l}{Real (Baseline)} & 210            & 1.59               & 87.60                                                                   & {\color[HTML]{F56B00} -}                                                           \\ \hline
\multicolumn{2}{l}{Inswapper}       & 20             & 2.45               & 50.00                                                                   & {\color[HTML]{F56B00} 72.42}                                                       \\\hline
\multicolumn{2}{l}{CanonSwap}       & 18             & 2.50               & 50.00                                                                   & {\color[HTML]{F56B00} 76.59}                                                       \\\hline
\multicolumn{2}{l}{DeepLiveCam}     & 19             & 2.58               & 42.10                                                                   & {\color[HTML]{F56B00} 72.44}                                                       \\\hline
\multicolumn{2}{l}{AvatarForcing}   & 141            & 2.62               & 47.52                                                                   & {\color[HTML]{F56B00} -}                                                           \\
\multicolumn{2}{r}{Real Audio}      & 23             & 2.57                & 43.48                                                                  & {\color[HTML]{F56B00} 74.02}                                                       \\
\multicolumn{2}{r}{ChatterBox}      & 26             & 2.75                & 43.75                                                                  & {\color[HTML]{F56B00} 78.19}                                                       \\
\multicolumn{2}{r}{FishAudio2}      & 32             & 2.86                & 38.24                                                                  & {\color[HTML]{F56B00} 79.20}                                                       \\
\multicolumn{2}{r}{OmniVoice}       & 33             & 2.42                & 55.17                                                                  & {\color[HTML]{F56B00} 79.50}                                                       \\
\multicolumn{2}{r}{Qwen3-TTS}       & 27             & 2.39                & 60.87                                                                  & {\color[HTML]{F56B00} 78.53}                                                       \\ \hline
\multicolumn{2}{l}{KeySync}         & 143            & 2.72               & 43.40                                                                   & {\color[HTML]{F56B00} -}                                                           \\
\multicolumn{2}{r}{Real Audio}      & 30             & 2.37                & 63.33                                                                  & {\color[HTML]{F56B00} 52.94}                                                       \\
\multicolumn{2}{r}{ChatterBox}      & 27             & 2.59                & 45.45                                                                  & {\color[HTML]{F56B00} 57.45}                                                       \\
\multicolumn{2}{r}{FishAudio2}      & 26             & 3.14                & 25.00                                                                    & {\color[HTML]{F56B00} 64.50}                                                       \\
\multicolumn{2}{r}{OmniVoice}       & 31             & 2.88                & 30.30                                                                  & {\color[HTML]{F56B00} 66.39}                                                       \\
\multicolumn{2}{r}{Qwen3-TTS}       & 29             & 2.57                & 53.33                                                                  & {\color[HTML]{F56B00} 60.78}                                                       \\\hline
\multicolumn{2}{l}{HunyuanPortrait} & 20             & 2.71                & 43.48                                                                  & {\color[HTML]{F56B00} 69.06}                                                       \\\hline
\multicolumn{2}{l}{MuseTalk}        & 121            & 2.73               & 43.80                                                                  & {\color[HTML]{F56B00} -}                                                           \\
\multicolumn{2}{r}{Real Audio}      & 30             & 2.67                & 50.00                                                                   & {\color[HTML]{F56B00} 75.19}                                                       \\
\multicolumn{2}{r}{ChatterBox}      & 22             & 2.59                & 50.00                                                                   & {\color[HTML]{F56B00} 77.51}                                                       \\
\multicolumn{2}{r}{FishAudio2}      & 19             & 2.58                & 47.37                                                                  & {\color[HTML]{F56B00} 77.34}                                                       \\
\multicolumn{2}{r}{OmniVoice}       & 21             & 3.33                & 14.29                                                                  & {\color[HTML]{F56B00} 81.97}                                                       \\
\multicolumn{2}{r}{Qwen3-TTS}       & 25              & 2.52                & 52.00                                                                   & {\color[HTML]{F56B00} 77.84}                                                       \\\hline
\multicolumn{2}{l}{LatentSync}      & 108 & 2.80 & 38.00 & {\color[HTML]{F56B00} -} \\

\multicolumn{2}{r}{Real Audio}      & 30  & 3.17 & 66.67 & {\color[HTML]{F56B00} 58.57} \\

\multicolumn{2}{r}{ChatterBox}      & 22  & 3.14 & 22.73 & {\color[HTML]{F56B00} 72.75} \\

\multicolumn{2}{r}{FishAudio2}      & 25  & 3.08 & 28.00 & {\color[HTML]{F56B00} 70.68} \\

\multicolumn{2}{r}{OmniVoice}       & 23  & 3.21 & 17.40 & {\color[HTML]{F56B00} 74.32} \\

\multicolumn{2}{r}{Qwen3-TTS}       & 8   & 2.25 & 62.50 & {\color[HTML]{F56B00} 71.88} \\ \hline                             
\multicolumn{2}{l}{HiFiFace}        & 27             & 2.85               & 37.00                                                                   & {\color[HTML]{F56B00} 65.56}                                                         \\\hline
\multicolumn{2}{l}{UniAvGen}        & 72             & 3.03               & 26.40                                                                   & {\color[HTML]{F56B00} 61.53}                                                        \\\hline
\multicolumn{2}{l}{PersonaLive}     & 137 & 3.48 & 10.90 & {\color[HTML]{F56B00} -} \\

\multicolumn{2}{r}{Real Audio}      & 26  & 3.07 & 26.92 & {\color[HTML]{F56B00} 78.79} \\

\multicolumn{2}{r}{ChatterBox}      & 28  & 3.57 & 3.57  & {\color[HTML]{F56B00} 81.02} \\

\multicolumn{2}{r}{FishAudio2}      & 30  & 3.63 & 6.67  & {\color[HTML]{F56B00} 80.04} \\

\multicolumn{2}{r}{OmniVoice}       & 27  & 3.37 & 11.11 & {\color[HTML]{F56B00} 81.23} \\

\multicolumn{2}{r}{Qwen3-TTS}       & 26  & 3.58 & 7.69  & {\color[HTML]{F56B00} 80.33}                                                   
\end{tabular}
\caption{Human evaluation results across manipulation methods, with audio manipulations breakdown. \textit{Mean Score} denotes the average human rating, while \textit{Deception Rate} reports the percentage of manipulated samples perceived as real (for real samples, the percentage correctly perceived as real). AUC reports the average performance of the best-performing audio-visual DeepFake detectors evaluated on the corresponding manipulation.}
\label{tab:appendix_broader_humaneval_breakdown}
\end{table*}

In Tables~\ref{tab:app_eval_audiobreakdown_p1} and~\ref{tab:app_eval_audiobreakdown_p2}, we report a fine-grained breakdown of audio-driven visual manipulations. While the main benchmark aggregates results across video synthesis pipelines by considering Fake and Real Audio, here we further break down performance by specific driving audio conditions: ChatterBox~\cite{chatterbox}, FishAudio2~\cite{fishaudio}, OmniVoice~\cite{omnivoice}, Qwen3-TTS~\cite{Qwen}, and Real Audio. This analysis evaluates whether detection difficulty is governed solely by the visual generation pipeline or modulated by the underlying speech synthesis engine.

The results reveal substantial performance variation across audio conditions, particularly for multimodal detectors. For example, the average audio-visual AUC on LatentSync~\cite{latentsync} ranges from 58.57 with Real Audio to 74.32 with OmniVoice; similarly, on KeySync~\cite{keysync}, performance ranges from 52.94 with Real Audio to 66.39 with OmniVoice. This sensitivity varies across visual engines: PersonaLive~\cite{personalive} remains consistently detectable across audio sources (78.79--81.23 AUC), whereas KeySync and LatentSync exhibit larger shifts depending on the driving audio. These results show that detection performance depends not only on the visual synthesis method, but also on the specific pairing of audio and visual generation techniques.

Modality-specific averages further highlight the role of joint audio-visual modeling. Across the full benchmark, audio-only models achieve an average AUC of 52.01, compared with 71.44 for video-only and 74.27 for multimodal architectures. However, the magnitude of this multimodal gain varies considerably across synthesis pairs, indicating that the usefulness of cross-modal cues depends on the artifact profile introduced by each generation pipeline.

\subsection{Human Evaluation Breakdown}
\label{app:human_eval_breakdown}

We extend the human evaluation reported in the main paper with a finer-grained analysis of audio-driven manipulations, as shown in Table~\ref{tab:appendix_broader_humaneval_breakdown}. Specifically, while the main paper reports results aggregated by video manipulation method, here we additionally break down KeySync, MuseTalk, LatentSync, AvatarForcing, and PersonaLive by audio manipulation. This allows us to examine how different voice-generation methods affect the perceived realism of the resulting audio-visual manipulations.

As in the main paper, real videos provide a reference baseline and are perceived as real in 87.6\% of the evaluations. Among manipulated samples, however, the human deception rate varies substantially both across video manipulation methods and across audio conditions. For example, KeySync achieves an overall deception rate of 43.4\%, but this ranges from 25.00\% with FishAudio2 to 63.33\% with real audio. Similarly, for LatentSync, the deception rate ranges from 17.40\% with OmniVoice to 66.67\% with real audio. These differences indicate that the perceived realism of an audio-driven manipulation depends not only on the visual generation method, but also on the audio used to drive it. Interestingly, the effect of a given audio-generation method is not consistent across visual methods, suggesting an interaction between the audio and visual components rather than a single voice-cloning method being systematically more deceptive.

For comparison with automated detection, we additionally report the average AUC of the best-performing audio-visual deepfake detectors for each condition. Human and machine difficulty do not necessarily follow the same trend. For instance, CanonSwap and InSwapper are perceived as real in 50.00\% of human evaluations, while the corresponding average detector AUCs remain 76.59 and 72.42, respectively. More generally, the results suggest that perceptual realism is insufficient to characterize detection difficulty: manipulations that are convincing to human observers may still contain discriminative audio-visual cues that are exploitable by automated detectors, while the relative difficulty of individual audio conditions can differ between humans and detection approaches.

\section{Qualitative Analysis}

\begin{figure*}[ht]
    \centering
    \includegraphics[scale=0.42]{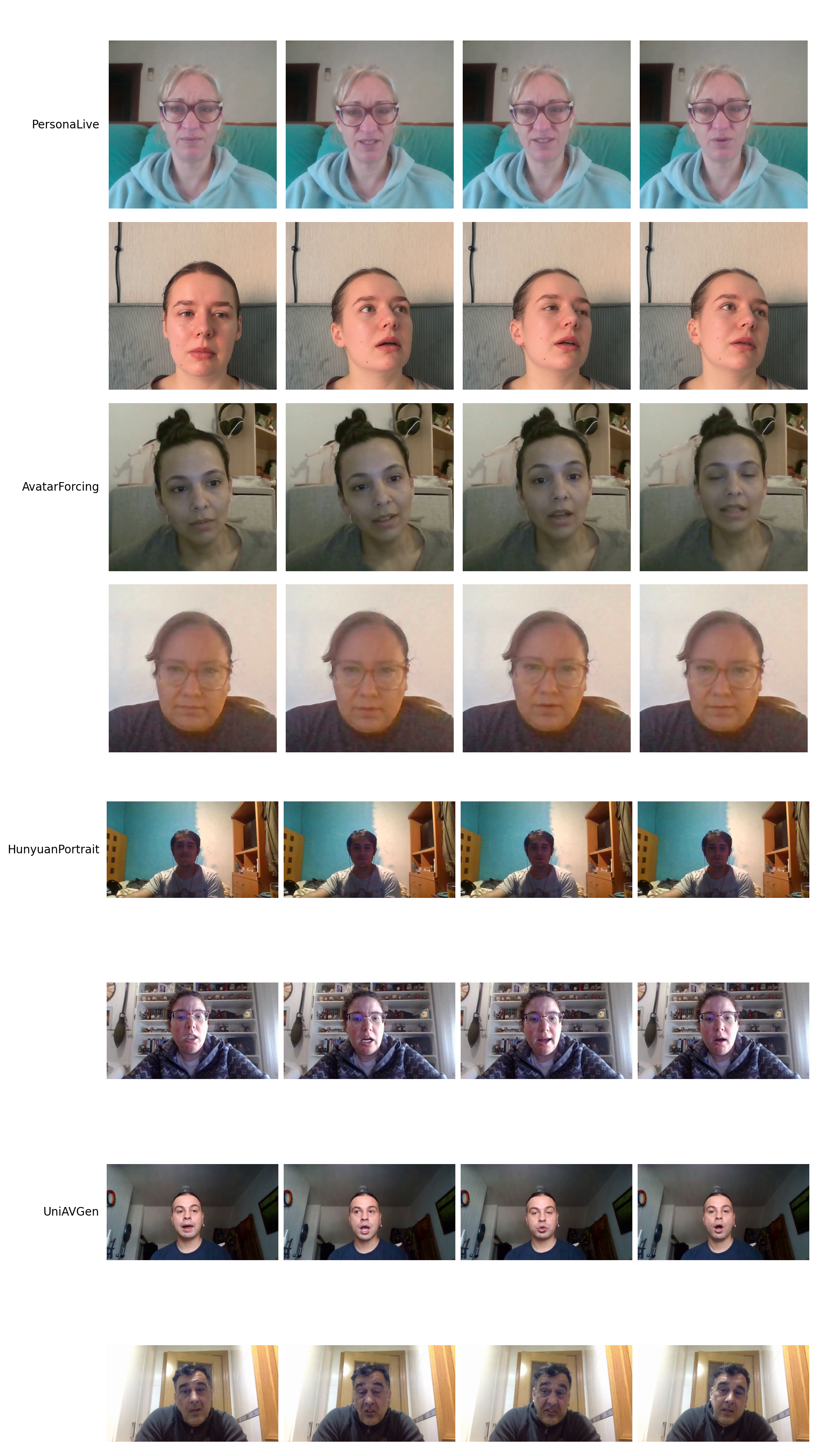}
    \caption{Qualitative examples of portrait animation manipulations in \ourname. Each row shows consecutive frames from a manipulated video generated with a different method. From top to bottom: PersonaLive~\cite{personalive}, AvatarForcing~\cite{avatarforcing}, HunyuanPortrait~\cite{hunyuanportrait}, and UniAVGen~\cite{uniavgen}.}
    \label{fig:avatar}
\end{figure*}

\begin{figure*}[ht]
    \centering
    \includegraphics[scale=0.45]{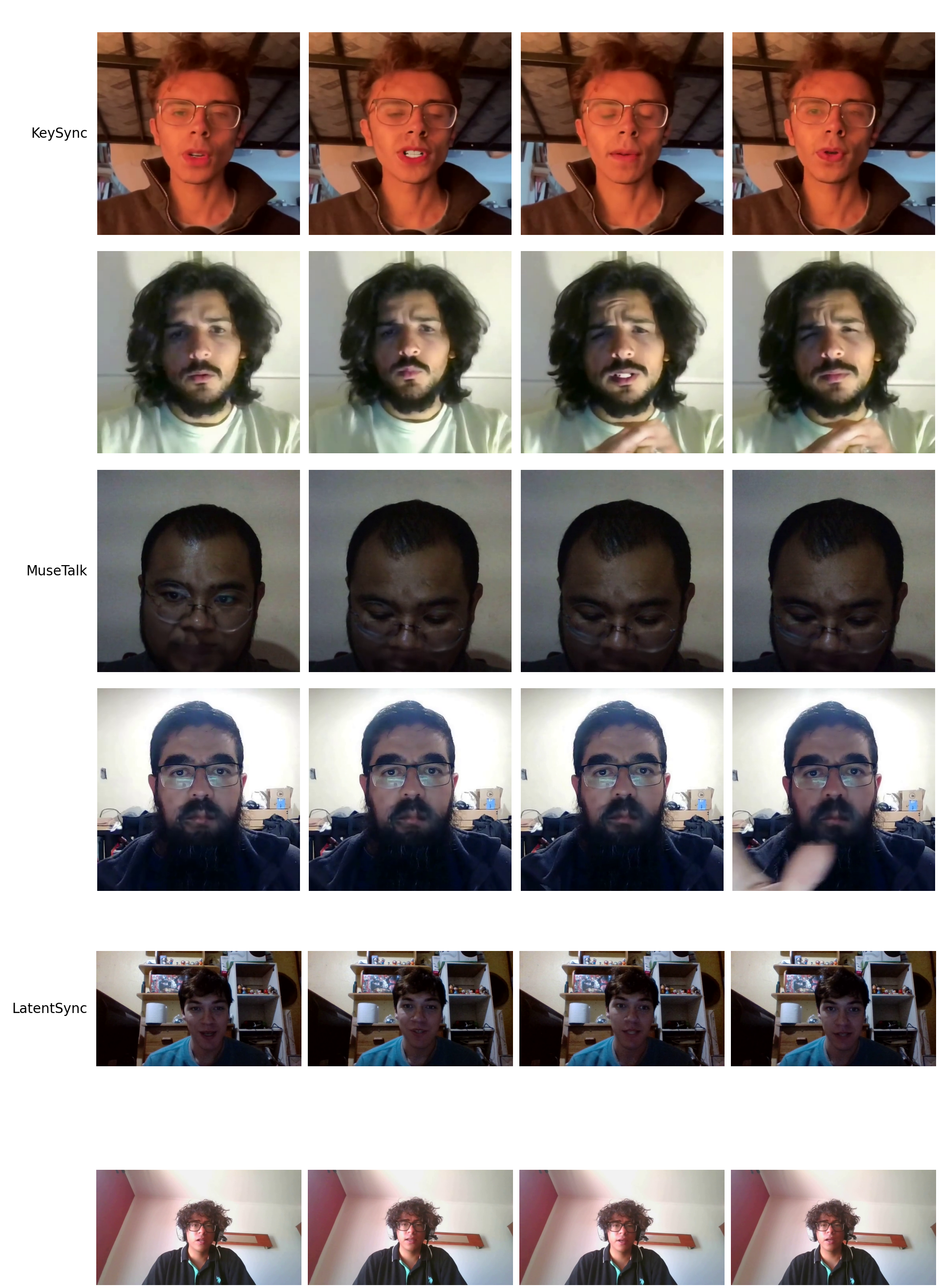}
    \caption{Qualitative examples of lip-synchronization manipulations in \ourname. Each row shows consecutive frames from a manipulated video generated with a different method. From top to bottom: KeySync~\cite{keysync}, MuseTalk~\cite{musetalk}, and LatentSync~\cite{latentsync}.}
    \label{fig:lipsynthesis}
\end{figure*}

\begin{figure*}[ht]
    \centering
    \includegraphics[scale=0.42]{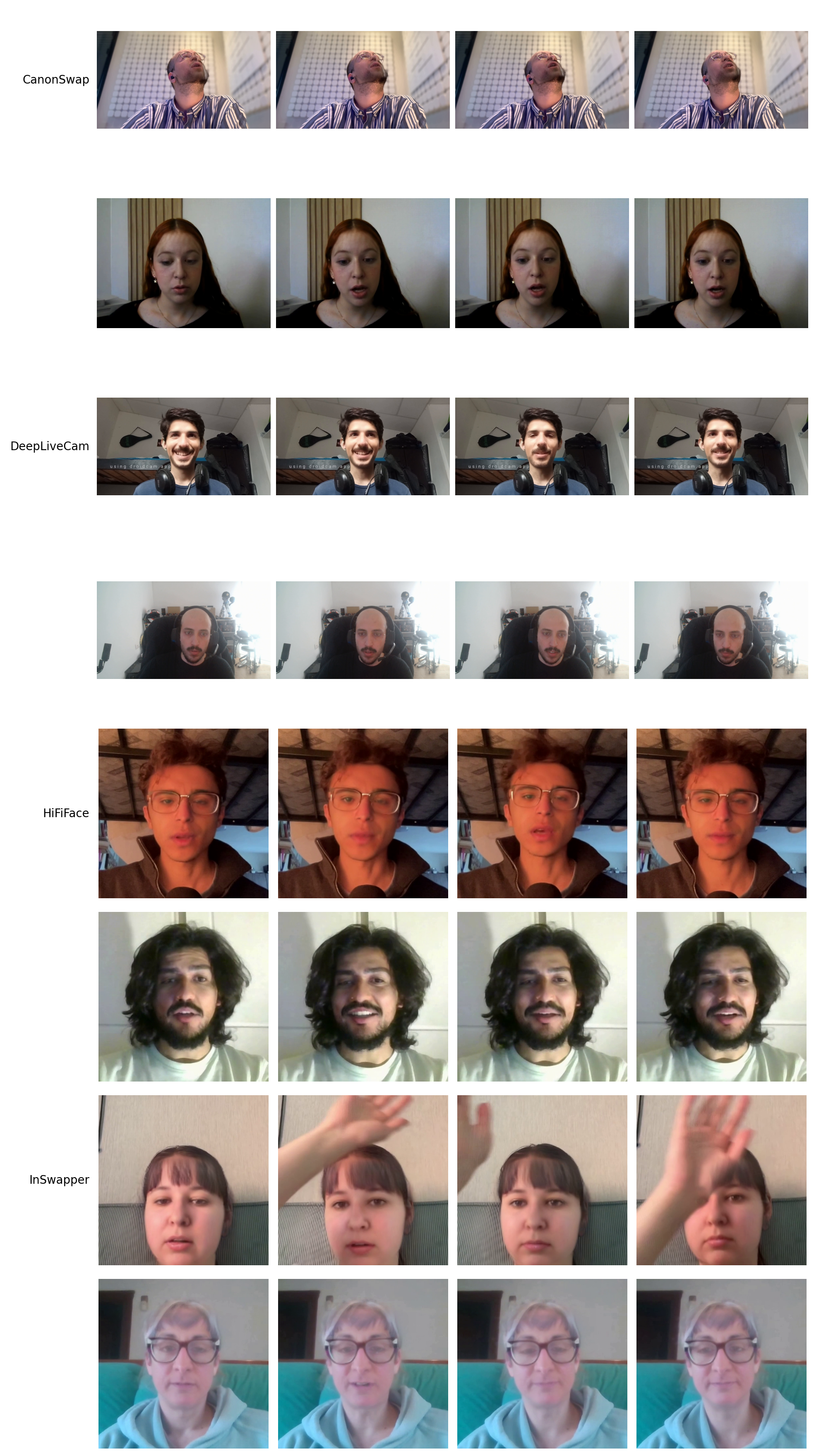}
    \caption{Qualitative examples of face-swapping manipulations in \ourname. Each row shows consecutive frames from a manipulated video generated with a different method. From top to bottom: CanonSwap~\cite{canonswap}, DeepLiveCam~\cite{deeplivecam}, HiFiFace~\cite{HifiFace}, and InSwapper~\cite{inswapper}.}
    \label{fig:faceswap}
\end{figure*}

To complement the quantitative evaluation, we qualitatively examine representative examples from \ourname across the three visual manipulation categories: portrait animation, lip synchronization, and face swapping. Figs.~\ref{fig:avatar}, \ref{fig:lipsynthesis}, and \ref{fig:faceswap} illustrate the diversity of outputs produced by the different generation pipelines included in the benchmark.
\newpage

\end{document}